# Plansformer: Generating Symbolic Plans using Transformers

Vishal Pallagani[1], Bharath Muppasani[1], Keerthiram Murugesan[2], Francesca Rossi[2], Lior Horesh[2], Biplav Srivastava[1], Francesco Fabiano[3], and Andrea Loreggia[4]

[1]*Artificial Intelligence Institute, University of South Carolina, USA*
[2]*IBM Research, USA*
[3]*University of Parma, Italy*
[4]*University of Brescia, Italy*

**Abstract**

Large Language Models (LLMs) have been the subject of active research, significantly advancing the field of Natural Language Processing (NLP). From BERT to BLOOM, LLMs have surpassed state-of-the-art results in various natural language tasks such as question answering, summarization, and text generation. Many ongoing efforts focus on understanding LLMs' capabilities, including their knowledge of the world, syntax, and semantics. However, extending the textual prowess of LLMs to symbolic reasoning has been slow and predominantly focused on tackling problems related to the mathematical field. In this paper, we explore the use of LLMs for automated planning - a branch of AI concerned with the realization of action sequences (plans) to achieve a goal, typically executed by intelligent agents, autonomous robots, and unmanned vehicles. We introduce Plansformer; an LLM fine-tuned on planning problems and capable of generating plans with favorable behavior in terms of correctness and length with reduced knowledge-engineering efforts. We also demonstrate the adaptability of Plansformer in solving different planning domains with varying complexities, owing to the transfer learning abilities of LLMs. For one configuration of Plansformer, we achieve ≈97% valid plans, out of which ≈95% are optimal for Towers of Hanoi - a puzzle-solving domain.

## 1 Introduction

Large Language Models (LLMs), based on transformer-based (neural) architecture [1, 2, 3, 4], have significantly advanced the field of Natural Language Processing (NLP). Their employment has grown dramatically in recent times [5], as researchers develop newer and bigger LLMs. From BERT to the most recent BLOOM, language models have surpassed state-of-the-art results in various natural language tasks. For example, PaLM [4], achieved breakthrough performance on a plethora of natural language tasks such as inference, question answering, and commonsense reasoning, and outperformed an average human performance on the BIG-bench benchmark.

Despite the textual prowess of LLMs, their significance has been limited in the domains that involve symbols. For example, domains with symbols such as mathematics [6, 7], and coding problems [8, 9] deliberates the failures of LLMs when it comes to handling symbols. In Automated Planning, [10] suggests that even state-of-the-art LLMs cannot reason with symbolic data and offer a new suite of benchmarks to test their reasoning capabilities. Recently, there has been a lot of interest in LLMs for code generation; for example, CodeT5 [11], CodeBERT [12], Codex [9], etc. In this paper, we propose to employ LLMs that are trained to generate code and repurpose them to generate valid plans. To advance the research in LLM-based automated planning, we create a training and test dataset for several planning domains. We use CodeT5 (base), a transformer-based code generation model that achieves state-of-the-art results in CodeXGlue, as the pre-trained LLM. We select CodeT5 due to its ability to generate goal-directed, sequential instruction and semantically meaningful program codes with syntactic and structural constraints. Then, we present, Plansformer, an LLM trained to generate symbolic plans of high quality in terms of correctness and length. Our experimental results indicate that the syntactic/symbolic knowledge learned from different programming languages in the CodeT5 model can be beneficial for the PDDL-based automated planning task. For example, in one configuration of Plansformer tested on a puzzle-solving domain, Towers of Hanoi - **hanoi**, our model was able to generate 97% valid plans, out of which 95% are shortest length plans. The results reveal a promising direction to harness LLMs for symbolic tasks such as planning.

In the remainder of the paper, we present preliminaries on automated planning and language models and then propose an LLM repurposed planner called *Plansformer*. Next, we present the experimental results comparing our approach with state-of-the-art planners and other large language models. Furthermore, we demonstrate the ability of Plansformer to adapt to other domains and discuss the relevance to instruction generation. We conclude with a discussion of the results and presentation of ongoing work.

## 2 Background

### 2.1 Automated Planning

Given the initial and goal states, alongside a set of legal actions, the objective of a planning agent is to devise a sequence of actions that advance the agent from the initial to the goal state. This paper adopts the Planning Domain Description Language (PDDL) [13, 14] notations. In PDDL, a planning environment is described in terms of objects in the world, predicates that describe relations between these objects, and actions that modify the world by manipulating these relations. The output plan consists of a series of time steps, each of which can have one or more instantiated actions with concurrency semantics [15]. A planner devises plans by searching in the space of states, where a state is a configuration of physical objects or partial plans. There is a single agent in the most basic formulation, called classical planning. The actions have unit cost, take constant time to execute, have deterministic effects, with the fully observable world, domain-specific conditions/constraints, and all goals have to be achieved [15]. In more sophisticated planning settings, many of these conditions are relaxed. There may be multiple agents, and the cost and duration of actions can be non-uniform. At the same time, its effects can be non-deterministic, the world can be partially observable, and the agent may maximize as many goals as it can achieve in a given time and resource budget.

### 2.2 Large Language Models and Symbolic Tasks

Large Language Models (LLM) such as BERT [16], RoBERTa [17], and GPT3 [3] are pre-trained on extensive unstructured knowledge from public data such as Wikipedia, Bookcorpus, and Commoncrawl, have shown impressive results in several NLP tasks. It demonstrated the ability to generalize to multiple tasks from question answering and machine translation to story generation and instruction following [18, 19]. LLMs have shown the ability to generate output in natural language [16, 20], adapt to novel tasks in a zero or few-shot approach [3, 21] and decode with constraints on output space [22, 23, 24]. Recent progress in LLMs has demonstrated the generation of structured output that requires precise syntactic/symbolic knowledge with structural constraints such as knowledge graphs [25], protein structure [26, 27], and programming languages [28]. As the LLMs collect the related knowledge necessary to solve an NLP task, [25] have shown that the LLMs are potential representations of the significant knowledge bases. In protein data [26, 27], LLMs generate the functional properties of the proteins by enforcing the structural constraints specific to protein science and determining the complex functional relationships in the protein binding. Code generation has recently become very popular in the LLM research community. Several models such as CodeBERT [12], Codex [9], and CodeT5 [11] have shown significant improvement in transfer from pre-trained models for natural language to structured codes. One of the key contributors to the success of these LLMs in code generation is fine-tuning the models on task-specific data. For instance, CodeXGlue [29], a benchmark dataset for code understanding and generation with sample codes from several programming languages, is used to fine-tune CodeBERT, CodeT5, and others. In this paper, we harness CodeT5 for further fine-tuning to the classical automated planning domain due to its ability to generate goal-directed, sequential instruction and semantically meaningful program codes with syntactic and structural constraints.

The closest prior art addressing the ability of LLMs to generate symbolic plans are the studies of [30, 10, 31]. [30] looks at different ways to generate plans using GPT-3 versions, a prominent generative LLM. [10] discusses different scenarios for generating plans, like finding a satisficing plan or adapting a previous one, and discusses encodings to generate plans and verify using a plan validator. [31] generates step-by-step instructions for a user-defined task using LLM prompting. All these studies require the (human-guided) mapping of a natural language-based sequence of instructions generated by the LLM to the admissible action in the planning domain as an additional step.

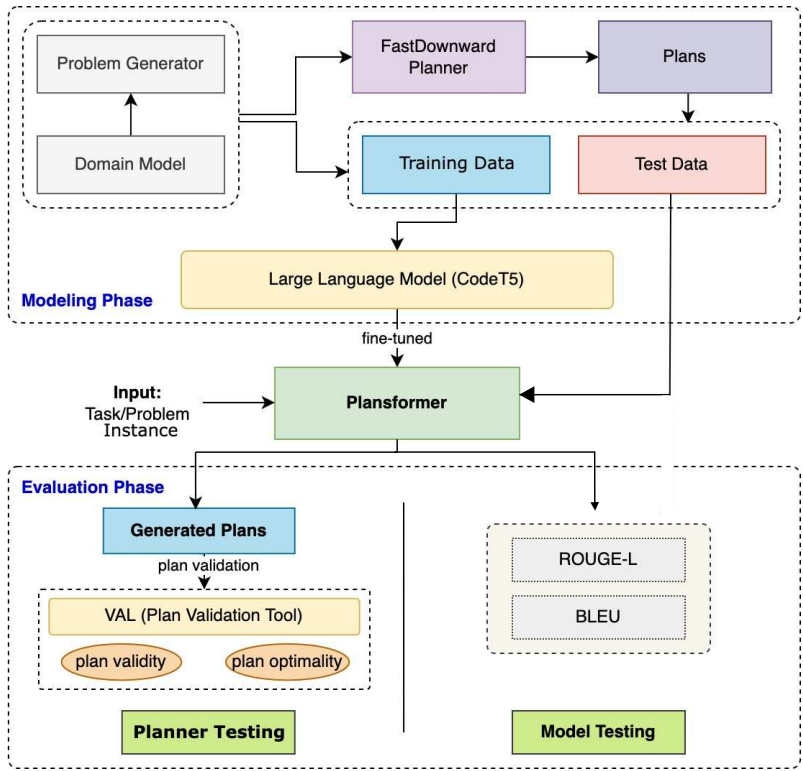

Figure 1: Plansformer Model Architecture showing modeling and evaluation phases. Modeling phase involves finetuning CodeT5 with data from planning domain. Evaluation phase shows both the planner and model testing.

## 3 Plansformer for Symbolic Plans

Figure 1 provides an illustrative overview of how we generate and test our planner, called Plansformer. The first phase, *modeling*, shows how we fine-tune the CodeT5 to address planning syntax and semantics. The second phase, *evaluation*, deals with assessing the competency of Plansformer as a model and as a planner. The key idea here is to utilize an LLM (CodeT5) pretrained on code generation and further train it on planning problem instances with corresponding valid plans. We evaluate its competence in generating valid plans (or almost valid plans for unseen planning problem instances) using two types of testing: 1) Model testing measures if Plansformer could generate meaningful responses (as in the test dataset), 2) Planner testing measures if the generated plans are valid/optimal (independently of whether they were the same plans as in the test dataset).

### 3.1 Modeling Phase

In the modeling phase, we first create a planning-based dataset for finetuning the CodeT5 to generate plans. The modeling phase of Figure 1 depicts the different modules employed.

#### 3.1.1 Planning dataset

We generate a PDDL-based dataset as a benchmark to finetune pretrained CodeT5 and facilitate further research at the intersection of LLMs and automated planning. We use the domain model (in PDDL) to generate corresponding valid problem files with varying complexities automatically. In this paper, we focus on four different classical planning domains, i.e., *Blocksworld, Towers of Hanoi, Grippers, Driverlog*.

| Task | Problem Instance | Plan |
|---|---|---|
| blocksworld | <GOAL> on b1 b2, on b2 b3, ontable b3, on b4 b1, clear b4<br><INIT> handempty, ontable b1, clear b1, on b2 b3, ontable b3, on b4 b2, clear b4<br><ACTION> pick-up<br>   <PRE> clear x, ontable x, handempty<br>   <EFFECT> not ontable x, not clear x, not handempty, holding x<br><ACTION> put-down<br>   <PRE> holding x<br>   <EFFECT> not holding x, clear x, handempty, ontable x<br><ACTION> stack<br>   <PRE> holding x, clear y<br>   <EFFECT> not holding x, not clear y, clear x, handempty, on x y<br><ACTION> unstack<br>   <PRE> on x y, clear x, handempty<br>   <EFFECT> holding x, clear y, not clear x, not handempty, not on x y | unstack b4 b2, put-down b4, pick-up b1, stack b1 b2, pick-up b4, stack b4 b1 |

Figure 2: Snapshot of one instance of the plan dataset for blocksworld domain.

*Blocksworld*, or **bw**, is a well-studied domain [32] with blocks placed on a table or arranged in vertical stacks. Here, one can alter the arrangement of the blocks with the available actions such as pick-up, put-down, stack, and unstack. We generate the problems with 2 to 5 block configurations.

*Towers of Hanoi*, or **hn**, consists of 3 pegs and multiple disks of varying diameters. Initially, all disks are placed in the first peg, and the end goal is to move the disks to the last peg. The only limitation to consider when moving the disks is that only a smaller disk can be placed on top of a bigger disk. Although the domain has only one action, the problem-solving is recursive [33]. Here, we generate the problems with configurations of 2 to 5 disks.

*Grippers*, or **gr** domain involves moving balls across rooms using robotic grippers. It has problems generated with configurations of 2 to 5 balls, 3 to 5 robots, and 2 to 4 rooms.

*Driverlog* or **dl** domain involves moving packages on trucks between locations driven by drivers. It has problems generated with configurations of 1 to 3 drivers, 1 to 3 trucks, 2 to 4 packages, and 3 to 6 locations.

Each planning domain explained above includes multiple problem instances. We generate the corresponding plans for each problem instance using FastDownward planner [34]. FastDownward is a classical planning system based on a heuristic search and offers different search algorithms such as causal graph heuristics and $A^*$ search. FastDownward can generate optimal plans with $A^*$ LM-Cut heuristic [35]. Hence, FastDownward can be regarded as a potent planner for generating a dataset of optimal plans.

We show the snapshot of the generated dataset in Figure 2, with more examples in Section 1 of supplementary material. Unlike the traditional planner which requires two different files - domain and problem (as pddl files, See Figure 2 in supplementary material), Plansformer reduces the knowledge-engineering efforts with a simplified input format that includes the problem instance coupled with a corresponding valid plan. The problem instance captures all the essential information in the domain and problem instance, such as the goal, the initial state, and the possible actions that can be taken in that domain. The generated dataset for each domain consists of 18,000 plans with different problem configurations. For training, we use 5-fold cross-validation with an 80%-20% split of the generated dataset for each domain. The average plan length (number of actions in the generated plan) for blocksworld is 9, gripper is 9, driverlog is 10, and hanoi is 12.

### 3.1.2 Tokenizer

We use a Byte-level BPE tokenizer, following the standard practice in LLMs, with a vocabulary size of 32,005. We add PDDL-specific tokens, namely, [GOAL], [INIT], [ACTION], [PRE], [EFFECT] to represent the goal state, initial state, possible actions with their associated preconditions and effects these actions cause in the environment respectively. We do not re-train a specific tokenizer for this task from scratch following the previous work [9], where GPT-3's tokenizer was reused to generate code.

### 3.1.3 Fine-tuning CodeT5

While there are many LLMs to select as a candidate for this work, we shortlist the models pre-trained on code generation to exploit the syntactic information in the programming languages implicitly captured in their weights. Although Codex [9], built using GPT-3, has reported the best performance in solving code-related tasks, its lack of public access led us to choose an equally competitive LLM: CodeT5 [11]. CodeT5 is a masked language model consisting of an encoder-decoder stack inspired by the transformer architecture [1]. It is capable of performing a wide range of tasks including code generation and understanding tasks. The generation tasks include code summarization, code generation, translation, and refinement. The understanding tasks include code defect detection and clone detection. CodeT5 is pretrained with example codes from eight programming languages - Python, Java, JavaScript, PHP, Ruby, Go, C, and C#. Its pre-training tasks include identifier awareness and bimodal generation, which optimizes code-to-code understanding. The CodeT5 model possesses several properties amenable to the planning domain, such as its ability to generate goal-directed, sequential instruction and semantically meaningful program codes with syntactic and structural constraints. With this pre-trained knowledge already encoded within CodeT5, we finetune it with 14400 samples (80% of the generated dataset) for each independent domain from the planning dataset. As a result of this finetuning, the weights of CodeT5 are updated to account for the task of plan generation. We give the planning problem instance as input to CodeT5's encoder and generate the intermediate features for the decoder of CodeT5 to output a plan.

## 3.2 Evaluation Phase

Plansformer is an LLM that ingests a new problem instance as input and outputs a plan for that problem instance. Therefore, to evaluate its competency, we must test its quality as a model and planner. The evaluation phase is described in the lower part of Figure 1, showing both testing phases.

### 3.2.1 Planner Testing

Unlike in natural language, symbolic plans have richer information content, inherently captured in their structure. Thus, we have an additional evaluation phase for plan validation to check how well Plansformer can mimic an automated planner. The sequence of actions generated by Plansformer must help an agent to navigate from the initial state to the goal state for a given problem instance. We call a generated plan *cost-optimal* [1] if it is the shortest possible among all other plans. Several metrics exist in the automated planning literature to evaluate a plan generated by Plansformer. In this paper, we consider *validity* and *optimality*. We evaluate the plan generated by Plansformer using a plan validation tool, called VAL [36], to check for its optimality and validity. VAL is an automatic validation tool for PDDL. VAL takes as input the task posed to Plansformer and the corresponding generated plan. It applies PDDL-based relaxation conditions to check for validity and optimality.

### 3.2.2 Model Testing.

It is typical to evaluate natural language tasks such as summarization or generation using metrics such as BLEU and ROUGE. Both BLEU and ROUGE are widely used metrics in NLP. In general, BLEU measures precision and helps understand how closely a machine translation (here, plan generated by Plansformer) is compared to a human translation (here, plan generated by an automated planner). On the other hand, ROUGE measures recall, i.e., how many of the words referenced in human summaries appeared in the summaries generated by the machine. In particular, we adopt ROUGE-L, which considers sentence-level structure similarity by identifying the longest co-occurring sequence n-grams. Although ROUGE and BLEU have no direct intuition in automated planning, we use these metrics to look at the task of plan generation from the perspective of LLMs. The evaluation based on these metrics provides us with an insight into the performance of Plansformer as a language model. In the next section, we evaluate Plansformer as a planner to give conclusive evidence on how well Plansformer generates the plans.

---

[1]We also refer to it as optimality interchangeably

| Models | ROUGE-L$_{recall}$ | ROUGE-L$_{precision}$ | ROUGE-L$_{fmeasure}$ | BLEU |
|---|---|---|---|---|
| Codex | 0.72 | 0.52 | 0.60 | 0.36 |
| GPT-2 | 0.04 | 0.14 | 0.06 | 0.07 |
| T5-base | 0.16 | 0.70 | 0.26 | 0.02 |
| CodeT5-base | 0.41 | 0.28 | 0.33 | 0.02 |
| Plansformer | **0.93** | **0.93** | **0.93** | **0.89** |
| Plansformer-bw | 0.97 | 0.99 | 0.98 | 0.90 |
| Plansformer-hn | 0.99 | 0.96 | 0.97 | 0.95 |
| Plansformer-gr | 0.94 | 0.94 | 0.94 | 0.92 |
| Plansformer-dl | 0.82 | 0.83 | 0.82 | 0.79 |

Table 1: Results of model testing (best performance in bold).

## 4 Experimental Results

In this section, we present the quantitative and qualitative results obtained using Plansformer to generate symbolic plans for multiple domains of varying complexities. We select a test-bed of 3,600 unique and unseen problem instances (20% of the dataset) for each domain for evaluating Plansformer. All the results reported in this paper are averaged over 5 randomly selected (80% − 20%) train-test splits. We report the results for the Plansformer variants by evaluating the corresponding test-bed. For example, Plansformer-bw's results are reported based on the performance results obtained on **bw** test-bed. We evaluate Plansformer using both model and planner testing to find its efficiency as a *language model* and a *planner*.

### 4.1 Is Plansformer a Good Model?

Plansformer has an encoder-decoder pair, where the encoder attends to tokens on either side of the masked word, whereas the decoder auto-regressively generates plans. Table 1 compares all the Plansformer models with other LLMs using the model evaluation metrics (ROUGE and BLEU). In this experiment, we consider the best-performing models from (bidirectional) masked language models (e.g., T5 [20]) and (unidirectional) causal language models (e.g., GPT-2 [21]) for the experiments. We present the actual plan generations from a few of these models in Figure 5 of supplementary material.

We report the performance of the baseline models averaged over the four planning domains. We also show the performance of Plansformer on individual domains (Plansformer-bw, Plansformer-hn, Plansformer-gr and Plansformer-dl). We observe that Plansformer performs best on all metrics, followed by Codex, with a significant ROUGE-L$_{recall}$ score. We believe that the performance gain from Codex compared to other baseline models is due to it's ability to relate the natural language understanding (a skill inherited from GPT-3) with code generation. It is interesting to see that CodeT5 performs poorly compared to Codex and Plansformer, demonstrating the advantages of the natural language understanding with code generation task on this evalutation metrics. We conclude that the models pre-trained with code-related tasks have an advantage over other models in plan generation task due to the similarities of PDDL with other programming languages. Despite with the best model testing metrics, We need to test Plansformer for plan validation to see its effectiveness as a planner.

### 4.2 Is Plansformer a Good Planner?

In this section, we report the results from the planner testing. We evaluate the generated plans for validity and optimality. We also report the average time taken to solve the problem instances. Table 2 shows the plan validation scores obtained by different models. We consider FastDownward [34] for plan validation to generate the ground truth plans. FastDownward planner generates a 100% valid and optimal plan for a given input (i.e., a combination of domain description and problem instance) when the landmark-cut heuristic is used within a standard A* search framework [35].

We can see that Blockworld **bw** domain achieved the highest performance gain via Plansformer-bw - generated 90.04%, out of which 88.44% are optimal. This better performance is analogous to the fact that **bw** is the easiest domain among the four domains. Although it is hard to find optimal plans, we can find a valid plan linear in the number of blocks to any problem instances by putting them down and picking from the table [32].

| Models | Valid Plans (%) | Invalid Plans | | Optimal Plans (%) | Avg. Time (sec) |
|---|---|---|---|---|---|
| | | Incomplete/Wrong (%) | Failed (%) | | |
| FastDownward (Ground Truth) | 100% | - | - | 100% | 10.28s |
| Codex | 0.15% | 0% | 99.85% | 0.15% | 1s |
| GPT-2 | 0% | 100% | 0% | 0% | 0.05s |
| T5-base | 0.25% | 82.7% | 17.3% | 0.25% | 0.47s |
| CodeT5-base | 0.6% | 99.4% | 0% | 0.6% | 0.68s |
| Plansformer | **83.64%** | 16.18% | 0.19% | **73.27%** | **0.06s** |
| Plansformer-bw | 90.04% | 9.94% | 0.02% | 88.44% | 0.05s |
| Plansformer-hn | 84.97% | 14.72% | 0.31% | 82.58% | 0.05s |
| Plansformer-gr | 82.97% | 16.61% | 0.42% | 69.47% | 0.06s |
| Plansformer-dl | 76.56% | 23.44% | 0% | 52.61% | 0.09s |

Table 2: Results of plan validation.

On a relatively more complex domain, i.e., **dl**, `Plansformer-dl` achieves 76.56% valid plans, out of which 52.61% are optimal. We notice a 20% difference between valid and optimal plans for **dl**, with an observation that the model can come up with completely new and valid action sequences, although may not optimal. We can see that the number of optimal plans generated reduces with the increasing complexity of the domains. We include both incomplete/wrong generations from the models and failed plans when reporting invalid plans. An incomplete/wrong generation is a partially correct ordering of action sequences, whereas a failed plan is an entire plan consists of impossible ordering of actions, not allowed by the domain definition. Figure 3 shows an example of incomplete generation and failed plans. All Plansformer models generate close to 0% failed plans for respective domains. It is a testament to Plansformer models' ability to understand and generate valid action sequences for a given domain.

**Failed Plans**

**Actual Plan:** unstack b2 b4, put-down b2, unstack b4 b1, put-down b4, unstack b1 b3, put-down b1, pick-up b2, stack b2 b1, pick-up b4, stack b4 b2

**Generated Plan:** unstack b2 b4, put-down b2, unstack b4 b1, put-down b4, unstack b1 b3, put-down b4, unstack b4 b1, put-down b4, unstack b1 b3, put-down b4, unstack b4 b2, put-down b4, unstack b2 b4, stack b2 b1, pick-up b4, stack b4 b2

**Incomplete Generations**

**Actual Plan:** unstack b1 b3, put-down b1, pick-up b4, stack b4 b1

**Generated Plan:** unstack b1 b3, put-down b1, pick-up

Figure 3: Different types of invalid plans generated by Plansformer on Blocksworld domain (Plansformer-bw).

Codex, the second best performing model according to ROUGE and BLEU scores, only generates 0.15% valid plans, emphasizing the need for a two-stage evaluation phase - where both model and generated plans are tested. We notice that the average time taken by Plansformer to completely solve the test-bed of problems is $\tilde{2}00x$ faster than the FastDownward, an automated planner that generated ground truth plans. Plansformer may offer an immense advantage in generating approximately correct plans in real-time applications. Interestingly, CodeT5, used to build Plansformer, takes considerable time to solve the same problem instances from the test bed. We believe that the Plansformer is faster since it generates valid and likely optimal plans shorter in length than usually long incoherent sequences generated by CodeT5, which tend to be time-consuming. The difference between plans generated by CodeT5 and Plansformer for the same problem is shown in Figure 4.

There have been very few relevant works that can be compared with Plansformer. Although some recent works such as [31] use LLMs to generate "plans", it is different from automated planning and is not symbolic in nature. These "plans" are step-by-step actions to perform a trivial everyday task such as *"Brush teeth"*. These methods use a user-constructed prompt to enable an LLM to generate appropriate steps from its prior knowledge. We believe the work by [10] is similar in spirit to ours, using a PDDL-based natural language prompt to obtain symbolic plans using GPT-3. The significant difference between their dataset and ours is the difference in the object names, for example, a block is named as *b1* in our dataset as opposed to *a* in [10]. This difference lets us evaluate Plansformer[2] on the dataset from [10] with randomized object names as opposed to our dataset introduced in Section 3.1.1. On this dataset, Plansformer generated 66% *valid plans*, whereas, GPT3 with PDDL-based natural language prompting generated only 0.6% valid plans. The significant difference in performance enables us to validate the advantage of our approach in generating valid plans despite randomized object names.

Plansformer, trained and tested on the same domain, displays superior performance both as a model and a planner. However, LLMs are also well known for transfer learning, i.e., a model trained for solving one domain can be re-purposed to solve other related domains. In the next section, we explore how Plansformer trained on one domain can be adapted to another.

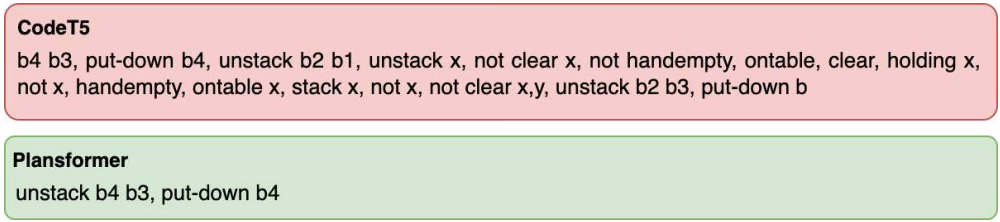

Figure 4: Long invalid generation from CodeT5 as opposed to a valid optimal plan from Plansformer-bw for the same problem.

## 4.3 Can Plansformer adapt to another domain?

The *base models*, i.e., Plansformer-x, where x can **bw, hn, gr, and dl**, cannot generate valid plans for other domains (i.e., `Plansformer-bw` on **hn, gr, and dl**) since each domain differs from the others in terms of action space, complexity, and solving. However, LLMs allow us to utilize the model trained in one domain to adapt to another using either making use of prompt conditioning or transfer learning with further fine-tuning on the problem instances from the new domain. We have seen from the previous work on prompt conditioning [10] that the performance of the model on an unseen domain is very sensitive to the manually-identified prompt. A small perturbation to the prompt can significantly affect the model's performance, and creating a perfect prompt requires understanding the inner workings of LLM on hand and trial and error. In recent years, researchers have started looking at automatic prompt generation [37], which we would like to explore in the future.

Instead of the prompt conditioning, we follow the transfer learning approach by finetuning Plansformer *base models* with problem instances from other domains to check the ability of Plansformer to adapt to new domains. For brevity, we demonstrate variants of `Plansformer-bw` models on three other domains. Figure 5 shows different `Plansformer-bw` models and their plan validation scores on respective test-bed from target domains. We report that the results for transfer learning setup of Plansformer *base models* convey the same insights as `Plansformer-bw` shown here and are presented in Section 4.2 of supplementary material.

We consider different numbers of problem instances for finetuning `Plansformer-bw` on a given domain to see how the performance of the model varies across the sample size. We use the model naming format to convey the details on the amount of problem instances used for finetuning the Plansformer base model, i.e., `bw-hn[500]` implies that we further finetune `Plansformer-bw` using 500 problem instances from **hn** and report the results. In Figure 5, we can see an overall increase in the number of valid plans for every testing domain as we increase the problem instances available for finetuning. We observe that the models fine-tuned with 2000, which is 14% of the training size of *base models*, achieves 50% of the valid plans recorded by `Plansformer-hn`, `Plansformer-gr`, and `Plansformer-dl`. Despite the complexity of these planning domains, we obtain

---

[2] We would like to note that Plansformer is trained only on the dataset introduced in Section 3.1.1 and is not trained/finetuned on the dataset in [10] for this experiment

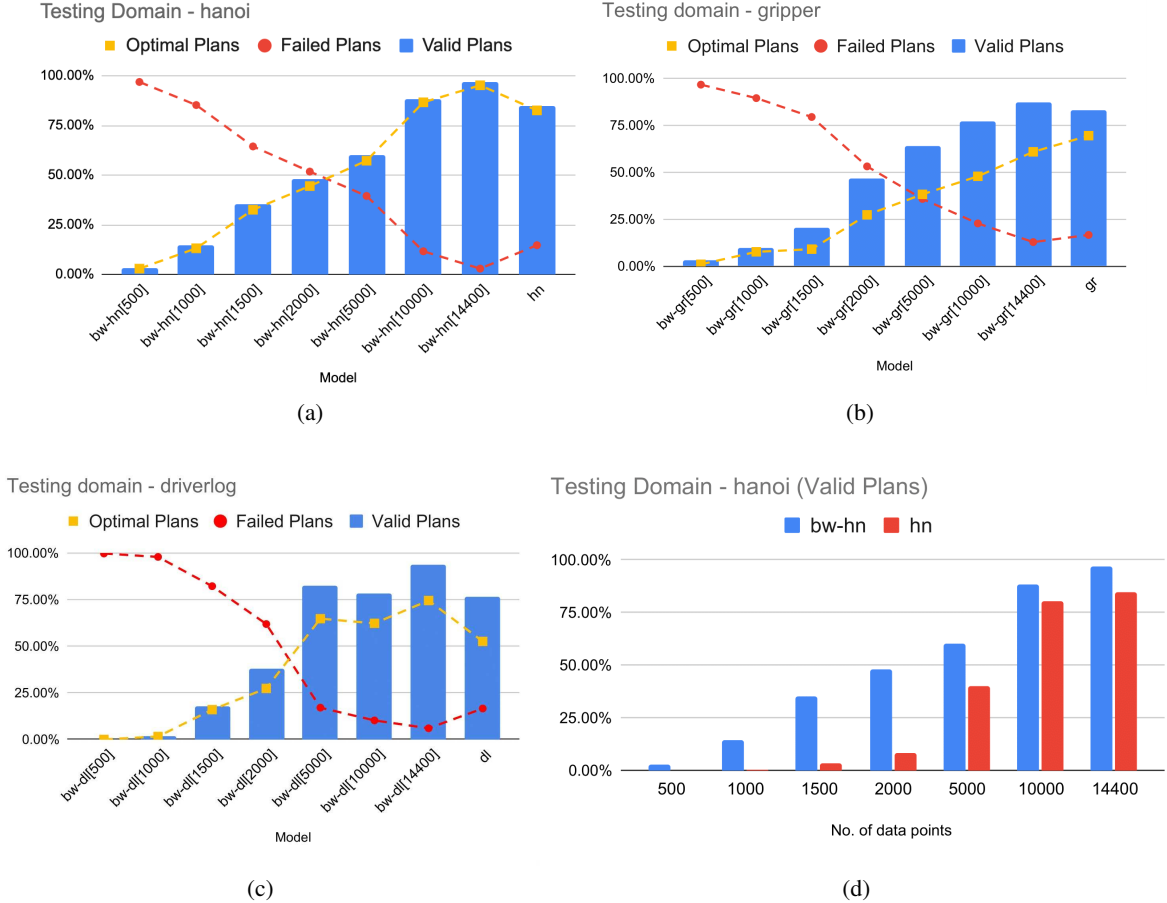

Figure 5: Plansformer-bw as the base model fine-tuned with and tested on (a) **hanoi** (b) **grippers** (c) **driverlog**, and (d) shows the comparison of valid plans generated by Plansformer-bw-hn derived models with Plansformer-hn trained using similar data points.

> 90% valid plans for all testing domains by increasing the finetuning samples to that of the training size of *base models*. `Plansformer-bw-hn[14400]` obtains the best performance among all models, by achieving 97.05% valid plans, out of which 95.22% are optimal. In Figure 5(d), we compare Plansformer-hn trained with different number of data points from **hn** domain against the Plansformer-bw (base model) finetuned on the same data points. We can see a clear advantage of the transfer learning capability in LLMs, as both the **bw** and **hn** domains have similar plan semantics. Similar trends can be seen for the other domains and the results are reported in Section 4.2 of supplementary material.

We notice that the failed plans decrease with additional problem instances used for finetuning. Using the same amount of problem instances as training (14, 400), we observe that the number of failed plans is less than that of the *base models* built for the respective domains. The number of optimal plans consistently increases with the number of problem instances in **hn** domain for finetuning. It is 13% more than `Plansformer-hn`, whereas we can see some variations for the other two domains. We also report the fine-tuning results of all possible Plansformer models and their performance in Figure 6 of the supplementary material. We have also trained a single Plansformer model on all the domains (multi-task setting) and found that the individual model has relatively comparable performance to that of 5 (See supplementary material Section 4.2 for more details).

## 5 Conclusions and Ongoing Work

In this paper, we have explored using LLMs to generate symbolic plans for multiple domains. We have taken an LLM tailored to code and trained it further over a set of planning problem instances and corresponding valid plans. We then tested the model's capability to generate plans for unseen planning problem instances, evaluating the correctness and length of such plans. Our approach is compared to an existing state-of-the-art planner, showing that our LLM-based planner, called Plansformer, can solve most instances with high quality both in terms of correctness and length while needing *much less time* to generate such plans. Although Plansformer shows promising results, this work is not done and can be extended in many directions. One can seek to eliminate invalid plans comprising both failed or incomplete plans, drive efficiency further, explore more planning domains, e.g., from competitions [38], and tackle other planning settings [10].

# References

[1] Ashish Vaswani, Noam Shazeer, Niki Parmar, Jakob Uszkoreit, Llion Jones, Aidan N Gomez, Lukasz Kaiser, and Illia Polosukhin. Attention is all you need. *Advances in neural information processing systems*, 30, 2017.

[2] Jacob Devlin, Ming-Wei Chang, Kenton Lee, and Kristina Toutanova. Bert: Pre-training of deep bidirectional transformers for language understanding. *arXiv preprint arXiv:1810.04805*, 2018.

[3] Tom Brown, Benjamin Mann, Nick Ryder, Melanie Subbiah, Jared D Kaplan, Prafulla Dhariwal, Arvind Neelakantan, Pranav Shyam, Girish Sastry, Amanda Askell, et al. Language models are few-shot learners. *Advances in neural information processing systems*, 33:1877–1901, 2020.

[4] Aakanksha Chowdhery, Sharan Narang, Jacob Devlin, Maarten Bosma, Gaurav Mishra, Adam Roberts, Paul Barham, Hyung Won Chung, Charles Sutton, Sebastian Gehrmann, et al. Palm: Scaling language modeling with pathways. *arXiv preprint arXiv:2204.02311*, 2022.

[5] Hang Li. Language models: Past, present, and future. *Commun. ACM*, 65(7):56–63, jun 2022.

[6] Dan Hendrycks, Collin Burns, Saurav Kadavath, Akul Arora, Steven Basart, Eric Tang, Dawn Song, and Jacob Steinhardt. Measuring mathematical problem solving with the math dataset. *arXiv preprint arXiv:2103.03874*, 2021.

[7] Karl Cobbe, Vineet Kosaraju, Mohammad Bavarian, Jacob Hilton, Reiichiro Nakano, Christopher Hesse, and John Schulman. Training verifiers to solve math word problems. *arXiv preprint arXiv:2110.14168*, 2021.

[8] Dan Hendrycks, Steven Basart, Saurav Kadavath, Mantas Mazeika, Akul Arora, Ethan Guo, Collin Burns, Samir Puranik, Horace He, Dawn Song, et al. Measuring coding challenge competence with apps. *arXiv preprint arXiv:2105.09938*, 2021.

[9] Mark Chen, Jerry Tworek, Heewoo Jun, Qiming Yuan, Henrique Ponde de Oliveira Pinto, Jared Kaplan, Harri Edwards, Yuri Burda, Nicholas Joseph, Greg Brockman, et al. Evaluating large language models trained on code. *arXiv preprint arXiv:2107.03374*, 2021.

[10] Karthik Valmeekam, Alberto Olmo, Sarath Sreedharan, and Subbarao Kambhampati. Large language models still can't plan (a benchmark for llms on planning and reasoning about change). *arXiv preprint arXiv:2206.10498*, 2022.

[11] Yue Wang, Weishi Wang, Shafiq Joty, and Steven CH Hoi. Codet5: Identifier-aware unified pre-trained encoder-decoder models for code understanding and generation. In *Proceedings of the 2021 Conference on Empirical Methods in Natural Language Processing*, pages 8696–8708, 2021.

[12] Zhangyin Feng, Daya Guo, Duyu Tang, Nan Duan, Xiaocheng Feng, Ming Gong, Linjun Shou, Bing Qin, Ting Liu, Daxin Jiang, et al. Codebert: A pre-trained model for programming and natural languages. In *Findings of the Association for Computational Linguistics: EMNLP 2020*, pages 1536–1547, 2020.

[13] Drew McDermott, Malik Ghallab, Craig Knoblock, David Wilkins, Anthony Barrett, Dave Christianson, Marc Friedman, Chung Kwok, Keith Golden, Scott Penberthy, David Smith, Ying Sun, and Daniel Weld. PDDL - the planning domain definition language. Technical report, Technical Report, 1998.

[14] M. Fox and D. Long. Pddl2.1: An extension to pddl for expressing temporal planning domains. *Journal of Artificial Intelligence Research*, 20:61–124, Dec 2003.

[15] Malik Ghallab, Dana Nau, and Paolo Traverso. *Automated Planning: Theory and Practice*. The Morgan Kaufmann Series in Artificial Intelligence. Morgan Kaufmann, Amsterdam, 2004.

[16] Thomas Wolf, Lysandre Debut, Victor Sanh, Julien Chaumond, Clement Delangue, Anthony Moi, Pierric Cistac, Tim Rault, Rémi Louf, Morgan Funtowicz, et al. Transformers: State-of-the-art natural language processing. In *Proceedings of the 2020 conference on empirical methods in natural language processing: system demonstrations*, pages 38–45, 2020.

[17] Yinhan Liu, Myle Ott, Naman Goyal, Jingfei Du, Mandar Joshi, Danqi Chen, Omer Levy, Mike Lewis, Luke Zettlemoyer, and Veselin Stoyanov. Roberta: A robustly optimized bert pretraining approach. *arXiv preprint arXiv:1907.11692*, 2019.

[18] Alex Wang, Amanpreet Singh, Julian Michael, Felix Hill, Omer Levy, and Samuel Bowman. Glue: A multitask benchmark and analysis platform for natural language understanding. In *Proceedings of the 2018 EMNLP Workshop BlackboxNLP: Analyzing and Interpreting Neural Networks for NLP*, pages 353–355, 2018.

[19] Alex Wang, Yada Pruksachatkun, Nikita Nangia, Amanpreet Singh, Julian Michael, Felix Hill, Omer Levy, and Samuel Bowman. Superglue: A stickier benchmark for general-purpose language understanding systems. *Advances in neural information processing systems*, 32, 2019.

[20] Colin Raffel, Noam Shazeer, Adam Roberts, Katherine Lee, Sharan Narang, Michael Matena, Yanqi Zhou, Wei Li, Peter J Liu, et al. Exploring the limits of transfer learning with a unified text-to-text transformer. *J. Mach. Learn. Res.*, 21(140):1–67, 2020.

[21] Alec Radford, Jeffrey Wu, Rewon Child, David Luan, Dario Amodei, Ilya Sutskever, et al. Language models are unsupervised multitask learners. *OpenAI blog*, 1(8):9, 2019.

[22] Chris Hokamp and Qun Liu. Lexically constrained decoding for sequence generation using grid beam search. In *Proceedings of the 55th Annual Meeting of the Association for Computational Linguistics (Volume 1: Long Papers)*, pages 1535–1546, 2017.

[23] Sean Welleck, Kianté Brantley, Hal Daumé Iii, and Kyunghyun Cho. Non-monotonic sequential text generation. In *International Conference on Machine Learning*, pages 6716–6726. PMLR, 2019.

[24] Sachin Kumar, Eric Malmi, Aliaksei Severyn, and Yulia Tsvetkov. Controlled text generation as continuous optimization with multiple constraints. *Advances in Neural Information Processing Systems*, 34:14542–14554, 2021.

[25] Fabio Petroni, Tim Rocktäschel, Sebastian Riedel, Patrick Lewis, Anton Bakhtin, Yuxiang Wu, and Alexander Miller. Language models as knowledge bases? In *Proceedings of the 2019 Conference on Empirical Methods in Natural Language Processing and the 9th International Joint Conference on Natural Language Processing (EMNLP-IJCNLP)*, pages 2463–2473, 2019.

[26] Serbulent Unsal, Heval Atas, Muammer Albayrak, Kemal Turhan, Aybar C Acar, and Tunca Doğan. Learning functional properties of proteins with language models. *Nature Machine Intelligence*, 4(3):227–245, 2022.

[27] Noelia Ferruz and Birte Höcker. Controllable protein design with language models. *Nature Machine Intelligence*, pages 1–12, 2022.

[28] Wasi Ahmad, Saikat Chakraborty, Baishakhi Ray, and Kai-Wei Chang. Unified pre-training for program understanding and generation. In *Proceedings of the 2021 Conference of the North American Chapter of the Association for Computational Linguistics: Human Language Technologies*, pages 2655–2668, 2021.

[29] Shuai Lu, Daya Guo, Shuo Ren, Junjie Huang, Alexey Svyatkovskiy, Ambrosio Blanco, Colin Clement, Dawn Drain, Daxin Jiang, Duyu Tang, et al. Codexglue: A machine learning benchmark dataset for code understanding and generation. In *Thirty-fifth Conference on Neural Information Processing Systems Datasets and Benchmarks Track (Round 1)*, 2021.

[30] Alberto Olmo Hernandez, Sarath Sreedharan, and Subbarao Kambhampati. Gpt3-to-plan: Extracting plans from text using GPT-3. *CoRR*, abs/2106.07131, 2021.

[31] Wenlong Huang, Pieter Abbeel, Deepak Pathak, and Igor Mordatch. Language models as zero-shot planners: Extracting actionable knowledge for embodied agents. *arXiv preprint arXiv:2201.07207*, 2022.

[32] Naresh Gupta and Dana S. Nau. Complexity results for blocks-world planning. In *In Proceedings of AAAI-91*, pages 629–633, 1991.

[33] C Gerety and P Cull. Time complexity of the towers of hanoi problem. *SIGACT News*, 18(1):80–87, mar 1986.

[34] Malte Helmert. The fast downward planning system. *Journal of Artificial Intelligence Research*, 26:191–246, 2006.

[35] Malte Helmert and Carmel Domshlak. Lm-cut: Optimal planning with the landmark-cut heuristic. *Seventh international planning competition (IPC 2011), deterministic part*, pages 103–105, 2011.

[36] Richard Howey, Derek Long, and Maria Fox. Val: Automatic plan validation, continuous effects and mixed initiative planning using pddl. In *16th IEEE International Conference on Tools with Artificial Intelligence*, pages 294–301. IEEE, 2004.

[37] Taylor Shin, Yasaman Razeghi, Robert L. Logan IV, Eric Wallace, and Sameer Singh. AutoPrompt: Eliciting knowledge from language models with automatically generated prompts. In *Empirical Methods in Natural Language Processing (EMNLP)*, 2020.

[38] ICAPS. International planning competitions at international conference on automated planning and scheduling (icaps). In *https://www.icaps-conference.org/competitions/*, 2022.

# Supplementary Material

# Contents

# 1 Dataset

In this section, we provide examples from the Planning dataset for each of the considered domain - **bw**, **hn**, **gr**, and **dl**. Figure 1 captures the different problem instances.

# 2 Planning vs Plansformer Input

We have talked about how a Plansformer brings about reduced knowledge engineering effort. In Figure 2a and Figure 2b, we show the input requirement for an Automated Planner for a **driverlog** problem configuration and Figure 3 shows corresponding input required by Plansformer for the same problem. An automated planner requires two files - (a) **domain.pddl**, and (b) **problem.pddl**. We reduce the knowledge engineering efforts in Plansformer by not requiring:

- explicit mention of *predicates* which are present in **domain.pddl** file.
- explicit mention of *objects* which are present in **problem.pddl** file.

We also have a conversion mechanism for Plansformer and Planning inputs, i.e., given a **domain.pddl** and **problem.pddl** files, we can convert them automatically to input required by Plansformer and vice versa.

| Task | Problem | Plan |
|---|---|---|
| blocksworld | <GOAL> on b1 b2, ontable b2, on b3 b1, on b4 b5, clear b4, on b5 b3<br><INIT> handempty, ontable b1, clear b1, ontable b2, clear b2, on b3 b5, clear b3, ontable b4, on b5 b4<br><ACTION> pick-up<br>　<PRE> clear x, ontable x, handempty<br>　<EFFECT> not ontable x, not clear x, not handempty, holding x<br><ACTION> put-down<br>　<PRE> holding x<br>　<EFFECT> not holding x, clear x, handempty, ontable x<br><ACTION> stack<br>　<PRE> holding x, clear y<br>　<EFFECT> not holding x, not clear y, clear x, handempty, on x y<br><ACTION> unstack<br>　<PRE> on x y, clear x, handempty<br>　<EFFECT> holding x, clear y, not clear x, not handempty, not on x y | unstack b4 b2, put-down b4, pick-up b1, stack b1 b2, pick-up b4, stack b4 b1 |

(a) Example from **blocksworld** dataset

| Task | Problem | Plan |
|---|---|---|
| hanoi | <GOAL> on d1 d2, clear d1, on d2 d4, on d3 peg2 , clear d3, on d4 peg1 , on d5 peg3 , clear d5<br><INIT> smaller peg1 d1, smaller peg1 d2, smaller peg1 d3, smaller peg1 d4, smaller peg1 d5, smaller peg2 d1, smaller peg2 d2, smaller peg2 d3, smaller peg2 d4, smaller peg2 d5, smaller peg3 d1, smaller peg3 d2, smaller peg3 d3, smaller peg3 d4, smaller peg3 d5, smaller d2 d1, smaller d3 d1, smaller d4 d1, smaller d5 d1, smaller d3 d2, smaller d4 d2, smaller d5 d2, smaller d4 d3, smaller d5 d3, smaller d5 d4, on d1 d2, clear d1, on d2 d5, on d3 peg1 , clear d3, on d4 peg2 , clear d4, on d5 peg3<br><ACTION> move<br>　<PRE> smaller to disc, on disc from, clear disc, clear to<br>　<EFFECT> clear from, on disc to, not on disc from, not clear to | move d1 d2 d3, move d2 d5 d4, move d1 d3 d2, move d3 peg1 d5, move d1 d2 peg1, move d2 d4 d3, move d1 peg1 d2, move d4 peg2 peg1, move d1 d2 peg2, move d2 d3 d4, move d1 peg2 d2, move d3 d5 peg2 |

(b) Example from **hanoi** dataset

| Task | Problem | Plan |
|---|---|---|
| grippers | <GOAL> at ball1 room3, at ball2 room2, at ball3 room3, at ball4 room2, at ball5 room3<br><INIT> at-robby robot1 room2, free robot1 lgripper1, free robot1 rgripper1, at-robby robot2 room1, free robot2 lgripper2, free robot2 rgripper2, at ball1 room3, at ball2 room1, at ball3 room1, at ball4 room1, at ball5 room3<br><ACTION> move<br>　<PRE> at-robby r from<br>　<EFFECT> at-robby r to, not at-robby r from<br><ACTION> pick<br>　<PRE> at obj room, at-robby r room, free r g<br>　<EFFECT> carry r obj g, not at obj room, not free r g<br><ACTION> drop<br>　<PRE> carry r obj g, at-robby r room<br>　<EFFECT> at obj room, free r g, not carry r obj g | pick robot2 ball2 room1 lgripper2, move robot1 room2 room1, pick robot1 ball3 room1 rgripper1, move robot1 room1 room3, drop robot1 ball3 room3 lgripper1, pick robot2 ball4 room1 rgripper2, move robot2 room1 room2, drop robot2 ball2 room2 lgripper2, drop robot2 ball4 room2 rgripper2 |

(c) Example from **grippers** dataset

| Task | Problem | Plan |
|---|---|---|
| driverlog | <GOAL> at package1 s2, at package2 s2, at package3 s3, at package4 s1, at package5 s3<br><INIT> at driver1 s2, at driver2 s4, at truck1 s4, empty truck1, at truck2 s3, empty truck2, at truck3 s3, empty truck3, link s1 s2, link s2 s1, link s1 s3, link s3 s1, link s2 s4, link s4 s2, link s3 s4, link s4 s3, link s1 s4, at package1 s2, at package2 s3, at package3 s1, at package4 s2, at package5 s4<br><ACTION> load-truck<br>　<PRE> at truck loc, at obj loc<br>　<EFFECT> not at obj loc, in obj truck<br><ACTION> unload-truck<br>　<PRE> at truck loc, in obj truck<br>　<EFFECT> not in obj truck, at obj loc<br><ACTION> board-truck<br>　<PRE> at truck loc, at driver loc, empty truck<br>　<EFFECT> not at driver loc, driving driver truck, not empty truck<br><ACTION> disembark-truck<br>　<PRE> at truck loc, driving driver truck<br>　<EFFECT> not driving driver truck, at driver loc, empty truck<br><ACTION> drive-truck<br>　<PRE> at truck loc-from, driving driver truck, link loc-from loc-to<br>　<EFFECT> not at truck loc-from, at truck loc-to<br><ACTION> walk<br>　<PRE> at driver loc-from, path loc-from loc-to<br>　<EFFECT> not at driver loc-from, at driver loc-to | board-truck driver2 truck1 s4, load-truck package5 truck1 s4, drive-truck truck1 s4 s1 driver2, load-truck package3 truck1 s1, drive-truck truck1 s1 s3 driver2, unload-truck package5 truck1 s3, unload-truck package3 truck1 s3, load-truck package2 truck1 s3, drive-truck truck1 s3 s1 driver2, drive-truck truck1 s1 s2 driver2, load-truck package4 truck1 s2, unload-truck package2 truck1 s2, drive-truck truck1 s2 s1 driver2, unload-truck package4 truck1 s1 |

(d) Example from **driverlog** dataset

Figure 1: Problem instances from four different planning domains

(a) Capturing **driverlog's** environment in `domain.pddl`

(b) Capturing the current state and desired goal of an object in **driverlog's** environment in `problem.pddl`

Figure 2: Files required to model a problem from **driverlog** in PDDL for execution by an Automated Planner

```
<GOAL> at package1 s1, at package2 s3, at package3 s3, at package4 s3
<INIT> at driver1 s3, at driver2 s3, at driver3 s3, at truck1 s3, empty truck1, at truck2 s3, empty truck2,
link s1 s2, link s2 s1, link s2 s3, link s3 s2, link s3 s1, link s1 s3, at package1 s3, at package2 s3, at
package3 s2, at package4 s1
<ACTION> load-truck
    <PRE> at truck loc, at obj loc
    <EFFECT> not at obj loc, in obj truck
<ACTION> unload-truck
    <PRE> at truck loc, in obj truck
    <EFFECT> not in obj truck, at obj loc
<ACTION> board-truck
    <PRE> at truck loc, at driver loc, empty truck
    <EFFECT> not at driver loc, driving driver truck, not empty truck
<ACTION> disembark-truck
    <PRE> at truck loc, driving driver truck
    <EFFECT> not driving driver truck, at driver loc, empty truck
<ACTION> drive-truck
    <PRE> at truck loc-from, driving driver truck, link loc-from loc-to
    <EFFECT> not at truck loc-from, at truck loc-to
<ACTION> walk
    <PRE> at driver loc-from, path loc-from loc-to
    <EFFECT> not at driver loc-from, at driver loc-to
```

Figure 3: Plansformer's input for the same problem defined in Figure 2a

## 3 Training Phase

In this section, we describe the hardware used for computation, training parameters and time taken by different models for training.

### 3.1 Hardware

We have used 9 (Dual P-100) 44 (Dual V100) GPU nodes for running our experiments. For training all models, we have made use of 24 cores of CPU run on 1 GPU node. Compute and GPU nodes have 128 GB of RAM and Big Data nodes have 1.5 TB RAM. All nodes have EDR infiniband (100 Gb/s) interconnects, and access to 1.4 PB of GPFS storage. The processor speed is 2.8 GHz.

### 3.2 Training Hyperparameters

Table 1 captures the hyperparameters set for training our models. For plan generation by all models apart from Codex, we have used beam search with number of beams set to 2, repetition penalty of 2.5, and length penalty set to 1.0. Codex doesn't have the functionality to change the parameters, thus, we have used it in the default setting. On the parameters constituting the considered models, Codex has 12 billion parameters, GPT-2 has 1.2 billion parameters, T5-base has 220 million parameters, and CodeT5-base has 8.35 million parameters.

| Hyperparameter | Value |
|---|---|
| Train Batch Size | 8 |
| Validation Batch Size | 8 |
| Train Epochs | 3 |
| Validation Epochs | 1 |
| Learning Rate | 1e-4 |
| Max Source Text Length | 512 |
| Max Target Text Length | 150 |

Table 1: Hyperparameters used for Training

### 3.3 Training Time

Figure 4 presents the training time taken by different Plansformer variations. Base models are Plansformer variants directly trained on each of the planning domains with CodeT5 as base. Derived models use a Plansformer base model as a starting point, and further pretrain on other domains. We can see a considerable drop in training time taken by derived models. It is also to be noted that these derived models outperform the base models when entire training data points are used.

## 4 Extended Results

This section adds additional qualitative and quantitative results that cover all the domains and model configurations tried and tested during our experimentation with Plansformer. In our initial testing phase, we have fine-tuned both T5 and CodeT5 with the same blocksworld dataset and hyperparameters and found that fine-tuned T5 gave 32% valid plans, whereas fine-tuned CodeT5 generated 90% valid plans. This is because CodeT5 has syntactically meaningful sequences for code-like structured inputs well defined as opposed to T5, which only deals with natural language. With this intuition that models pre-trained on code have an advantage for plan generation, we proceeded with the choice of using CodeT5 for all our experiments.

### 4.1 Qualitative Analysis

Figure 5 shows the output generations obtained by different models under study for a problem instance from each of the planning domains. When reporting Plansformer results, we take the *base models* corresponding to the domain being tested.

### 4.2 Quantitative Analysis

Figure 6 captures the performance of all models in their ability to generate plans for multiple domains.

Figures 7 to 15 represent the performance of different base models fine-tuned and tested on other domains in graphical manner. Additionally, we also wanted to check

Plansformer's capability in plan generation when the plan length of test set is considerably larger than that of the train set. For this purpose, we have created a train set for blocksworld consisting of 2,3 block configurations and a test set with 4,5 block configurations. The test set consists of 100 total instances, 50 from 4 block configuration and 50 from 5 block configuration. The average plan length for the train set and test set is 4 and 10 respectively. Plansformer trained on 2,3 block configurations was able to generate 64% valid plans on problem instances from the test set, showing that our approach can generate valid plans even if the plan length for test set >>> train set. We were able to achieve 64% valid plans in this experimentation as the train set consists of only 162 data points (all possible 2,3 block configurations) as compared to 87% obtained by Plansformer-bw (trained on 14,400 data points) on the same test set.

### 4.2.1 Transfer Learning

We have reported the advantages of transfer learning when using Plansformer-bw as the base model and further fine-tuning it on **hn** for all the varying data points. Figures 16 and 17 show a similar trend for the domains **dl** and **gr** on further fine-tuning of the Plansformer-bw base model.

### 4.2.2 Multi-domain Plan Generation using a Single Model

We have additionally trained a single Plansformer model on all the training data points belonging to all the four domains - **bw, hn, dl,** and **gr**. Each domain consists of 14,400 training data points, thus, the single Plansformer model is trained on a total of 57,600 data points. The obtained single Plansformer model is tested on the validation datasets corresponding to all the four domains. Each dataset consists of 3000 problem instances. Table 2 reports the performance obtained by the single Plansformer model on all four domains in terms of plan validation. We observe that the single model is able to perform relatively comparable to the *base models*. The single Plansformer trained on all models outperforms Plansformer-bw by 5.79% (in terms of valid plans) and has around a 4.5% to 13.5% decrease in valid plans for the other three domains.

| Test Domain | Valid Plans | Cost-Optimal Plans | Invalid Plans |
|---|---|---|---|
| **bw** | 95.83% | 93.75% | Failed = 4.17%, Incomplete = 0% |
| **hn** | 79.25% | 76.72% | Failed = 2.34%, Incomplete = 18.41% |
| **gr** | 78.44% | 50.61% | Failed = 15.03%, Incomplete = 6.53% |
| **dl** | 63.03% | 55.25% | Failed = 2.81%, Incomplete = 34.16% |

Table 2: Results of TREAT System

## 5 Plansformer Architecture

Figure 18 shows the layer-wise architecture that makes up Plansformer. All these layers are updated during the fine-tuning process. We inherit this architecture from CodeT5, and do not freeze any layers during our fine-tuning process involved for constructing Plansformer.

| | Models | Training Time |
|---|---|---|
| Base Models | plansformer-bw | 1hr 1min |
| | plansformer-hn | 1hr 10mins |
| | plansformer-gr | 1hr 12mins |
| | plansformer-dl | 1hr 15mins |
| Derived Models | plansformer-bw-dl[500] | 3 mins |
| | plansformer-bw-dl[1000] | 5 mins |
| | plansformer-bw-dl[1500] | 7 mins |
| | plansformer-bw-dl[2000] | 13 mins |
| | plansformer-bw-dl[5000] | 21 mins |
| | plansformer-bw-dl[10000] | 38 mins |
| | plansformer-bw-dl[14400] | 46 mins |
| | plansformer-bw-gr[500] | 3 mins |
| | plansformer-bw-gr[1000] | 5 mins |
| | plansformer-bw-gr[1500] | 7 mins |
| | plansformer-bw-gr[2000] | 13 mins |
| | plansformer-bw-gr[5000] | 19 mins |
| | plansformer-bw-gr[10000] | 37 mins |
| | plansformer-bw-gr[14400] | 44 mins |
| | plansformer-bw-hn[500] | 2 mins |
| | plansformer-bw-hn[1000] | 3 mins |
| | plansformer-bw-hn[1500] | 5 mins |
| | plansformer-bw-hn[2000] | 9 mins |
| | plansformer-bw-hn[5000] | 18 mins |
| | plansformer-bw-hn[10000] | 23 mins |
| | plansformer-bw-hn[14400] | 32 mins |
| | plansformer-hn-bw[500] | 2 mins |
| | plansformer-hn-bw[1000] | 2 mins |
| | plansformer-hn-bw[1500] | 3 mins |
| | plansformer-hn-bw[2000] | 5 mins |
| | plansformer-hn-bw[5000] | 8 mins |
| | plansformer-hn-bw[10000] | 12 mins |
| | plansformer-hn-bw[14400] | 17 mins |
| | plansformer-hn-gr[500] | 2 mins |
| | plansformer-hn-gr[1000] | 3 mins |
| | plansformer-hn-gr[1500] | 5 mins |
| | plansformer-hn-gr[2000] | 9 mins |
| | plansformer-hn-gr[5000] | 18 mins |
| | plansformer-hn-gr[10000] | 23 mins |
| | plansformer-hn-gr[14400] | 29 mins |
| | plansformer-hn-dl[500] | 1 min |
| | plansformer-hn-dl[1000] | 5 mins |
| | plansformer-hn-dl[1500] | 5 mins |
| | plansformer-hn-dl[2000] | 12 mins |
| | plansformer-hn-dl[5000] | 19 mins |
| | plansformer-hn-dl[10000] | 33 mins |
| | plansformer-hn-dl[14400] | 39 mins |
| | plansformer-gr-bw[500] | 2 mins |
| | plansformer-gr-bw[1000] | 2 mins |
| | plansformer-gr-bw[1500] | 3 mins |
| | plansformer-gr-bw[2000] | 5 mins |
| | plansformer-gr-bw[5000] | 8 mins |
| | plansformer-gr-bw[10000] | 14 mins |
| | plansformer-gr-bw[14400] | 18 mins |
| | plansformer-gr-hn[500] | 1 min |
| | plansformer-gr-hn[1000] | 2 mins |
| | plansformer-gr-hn[1500] | 2 mins |
| | plansformer-gr-hn[2000] | 5 mins |
| | plansformer-gr-hn[5000] | 10 mins |
| | plansformer-gr-hn[10000] | 13 mins |
| | plansformer-gr-hn[14400] | 16 mins |
| | plansformer-gr-dl[500] | 3 mins |
| | plansformer-gr-dl[1000] | 5 mins |
| | plansformer-gr-dl[1500] | 7 mins |
| | plansformer-gr-dl[2000] | 13 mins |
| | plansformer-gr-dl[5000] | 21 mins |
| | plansformer-gr-dl[10000] | 38 mins |
| | plansformer-gr-dl[14400] | 46 mins |
| | plansformer-dl-bw[500] | 1 min |
| | plansformer-dl-bw[1000] | 1 min |
| | plansformer-dl-bw[1500] | 2 mins |
| | plansformer-dl-bw[2000] | 2 mins |
| | plansformer-dl-bw[5000] | 4 mins |
| | plansformer-dl-bw[10000] | 7 mins |
| | plansformer-dl-bw[14400] | 13 mins |
| | plansformer-dl-hn[500] | 1 min |
| | plansformer-dl-hn[1000] | 2 mins |
| | plansformer-dl-hn[1500] | 2 mins |
| | plansformer-dl-hn[2000] | 5 mins |
| | plansformer-dl-hn[5000] | 8 mins |
| | plansformer-dl-hn[10000] | 12 mins |
| | plansformer-dl-hn[14400] | 16 mins |
| | plansformer-dl-gr[500] | 2 mins |
| | plansformer-dl-gr[1000] | 5 mins |
| | plansformer-dl-gr[1500] | 7 mins |
| | plansformer-dl-gr[2000] | 12 mins |
| | plansformer-dl-gr[5000] | 21 mins |
| | plansformer-dl-gr[10000] | 29 mins |
| | plansformer-dl-gr[14400] | 37 mins |

Figure 4: Training time of different Plansformer variants

Figure 5: Output generations from different models for planning problem instances

| | Test Domain - (tested using 3600 problems for each domain) | | | | | | | | | | | |
|---|---|---|---|---|---|---|---|---|---|---|---|---|
| | blocksworld (bw) | | | hanoi (hn) | | | gripper (gr) | | | driverlog (dl) | | |
| | Valid Plans | Invalid Plans | Cost-Optimal Plans | Valid Plans | Invalid Plans | Cost-Optimal Plans | Valid Plans | Invalid Plans | Cost-Optimal Plans | Valid Plans | Invalid Plans | Cost-Optimal Plans |
| plansformer-bw | 90.04% | Failed = 9.94%, Incomplete = 0.02% | 88.44% | 0.00% | Failed = 100%, Incomplete = 0% | 0.00% | 0.00% | Failed = 100%, Incomplete = 0% | 0.00% | 0.00% | Failed = 0%, Incomplete = 100% | 0.00% |
| plansformer-hn | 0.00% | Failed = 2.58%, Incomplete = 97.42% | 0.00% | 84.97% | Failed = 14.72%, Incomplete = 0.31% | 82.58% | 0.00% | Failed = 1.14%, Incomplete = 98.86% | 0.00% | 0.00% | Failed = 0%, Incomplete = 100% | 0.00% |
| plansformer-gr | 0.00% | Failed = 1.75%, Incomplete = 98.25% | 0.00% | 0.00% | Failed = 8.83%, Incomplete = 91.16% | 0.00% | 82.97% | Failed = 16.61%, Incomplete = 0.42% | 69.47% | 0.00% | Failed = 0%, Incomplete = 100% | 0.00% |
| plansformer-dl | 0.00% | Failed = 0%, Incomplete = 100% | 0.00% | 0.00% | Failed = 0%, Incomplete = 100% | 0.00% | 0.00% | Failed = 100%, Incomplete = 0% | 0.00% | 76.56% | Failed = 23.44%, Incomplete = 0% | 52.61% |
| plansformer-bw-dl[500] | 68.28% | Failed = 31.72%, Incomplete = 0% | 66.56% | 0.00% | Failed = 0%, Incomplete = 100% | 0.00% | 0.00% | Failed = 23.17%, Incomplete = 76.83% | 0.00% | 0.00% | Failed = 99.81%, Incomplete = 0.19% | 0.00% |
| plansformer-bw-dl[1000] | 59.22% | Failed = 40.78%, Incomplete = 0% | 58.44% | 0.00% | Failed = 0.03%, Incomplete = 99.97% | 0.00% | 0.00% | Failed = 0%, Incomplete = 100% | 0.00% | 1.92% | Failed = 97.97%, Incomplete = 0.11% | 1.58% |
| plansformer-bw-dl[1500] | 61.67% | Failed = 38.33%, Incomplete = 0% | 60.86% | 0.00% | Failed = 0.17%, Incomplete = 99.83% | 0.00% | 0.00% | Failed = 83.94%, Incomplete = 16.06% | 0.00% | 17.57% | Failed = 82.26%, Incomplete = 0.17% | 15.99% |
| plansformer-bw-dl[2000] | 49.57% | Failed = 50.43%, Incomplete = 0% | 48.66% | 0.00% | Failed = 0.17%, Incomplete = 99.83% | 0.00% | 0.00% | Failed = 83.94%, Incomplete = 16.06% | 0.00% | 37.81% | Failed = 61.94%, Incomplete = 0.25% | 27.39% |
| plansformer-bw-dl[5000] | 33.44% | Failed = 56.69%, Incomplete = 9.86% | 32.50% | 0.00% | Failed = 0.14%, Incomplete = 99.86% | 0.00% | 0.00% | Failed = 94.69%, Incomplete = 5.31% | 0.00% | 82.72% | Failed = 17.06%, Incomplete = 0.22% | 64.81% |
| plansformer-bw-dl[10000] | 3.25% | Failed = 10.64%, Incomplete = 86.11% | 3.25% | 0.00% | Failed = 0%, Incomplete = 100% | 0.00% | 0.00% | Failed = 100%, Incomplete = 0% | 0.00% | 78.56% | Failed = 10.22%, Incomplete = 11.22% | 62.33% |
| plansformer-bw-dl[14400] | 0.00% | Failed = 1%, Incomplete = 99% | 0.00% | 0.00% | Failed = 2.56%, Incomplete = 97.44% | 0.00% | 0.00% | Failed = 100%, Incomplete = 0% | 0.00% | 93.75% | Failed = 6%, Incomplete = 0.25% | 74.56% |
| plansformer-bw-gr[500] | 32.33% | Failed = 40.03%, Incomplete = 27.64% | 32.14% | 0.00% | Failed = 4.89%, Incomplete = 95.11% | 0.00% | 2.86% | Failed = 96.64%, Incomplete = 0.5% | 0.97% | 0.00% | Failed = 0%, Incomplete = 100% | 0.00% |
| plansformer-bw-gr[1000] | 17.03% | Failed = 13.81%, Incomplete = 69.17% | 17.00% | 0.00% | Failed = 0.67%, Incomplete = 99.33% | 0.00% | 9.42% | Failed = 89.47%, Incomplete = 1.11% | 7.61% | 0.00% | Failed = 0%, Incomplete = 100% | 0.00% |
| plansformer-bw-gr[1500] | 6.39% | Failed = 10.78%, Incomplete = 82.83% | 6.28% | 0.00% | Failed = 9.72%, Incomplete = 90.28% | 0.00% | 20.53% | Failed = 79.42%, Incomplete = 0.05 | 9.03% | 0.00% | Failed = 0%, Incomplete = 100% | 0.00% |

Trained using 14,400 problem instances for each domain

Figure 6: Plan Validation metrics for all Plansformer variants

| | | | | | | |
|---|---|---|---|---|---|---|
| plansformer-bw-gr[2000] | Failed = 18.42%, Incomplete = 72.28% | 9.31% | Failed = 0.92%, Incomplete = 99.08% | 0.00% | Failed = 73.11%, Incomplete = 0.33% | 26.56% | Failed = 0%, Incomplete = 100% | 0.00% |
| plansformer-bw-gr[5000] | Failed = 3.64%, Incomplete = 95.08% | 1.28% | Failed = 39.03%, Incomplete = 60.97% | 0.00% | Failed = 35.78%, Incomplete = 0% | 64.22% | Failed = 0%, Incomplete = 100% | 0.00% |
| plansformer-bw-gr[10000] | Failed = 1.56%, Incomplete = 98.44% | 0.00% | Failed = 6.06%, Incomplete = 93.94% | 0.00% | Failed = 22.75%, Incomplete = 0.03% | 77.22% | Failed = 0%, Incomplete = 100% | 0.00% |
| plansformer-bw-gr[14400] | Failed = 0%, Incomplete = 100% | 0.00% | Failed = 2.03%, Incomplete = 97.97% | 0.00% | Failed = 12.80%, Incomplete = 0.03% | 87.17% | Failed = 0%, Incomplete = 100% | 0.00% |
| plansformer-bw-hn[500] | Failed = 58.33%, Incomplete = 0.81% | 40.86% | Failed = 96.89%, Incomplete = 0% | 3.11% | Failed = 4.92%, Incomplete = 95.08% | 0.00% | Failed = 0%, Incomplete = 100% | 0.00% |
| plansformer-bw-hn[1000] | Failed = 57.81%, Incomplete = 4.39% | 37.81% | Failed = 85.28%, Incomplete = 0% | 14.72% | Failed = 4.92%, Incomplete = 95.08% | 0.00% | Failed = 0%, Incomplete = 100% | 0.00% |
| plansformer-bw-hn[1500] | Failed = 60.53%, Incomplete = 7.83% | 31.64% | Failed = 64.44%, Incomplete = 0.28% | 35.28% | Failed = 14.58%, Incomplete = 85.42% | 0.00% | Failed = 0%, Incomplete = 100% | 0.00% |
| plansformer-bw-hn[2000] | Failed = 56.22%, Incomplete = 25.81% | 17.97% | Failed = 51.78%, Incomplete = 0.03% | 48.19% | Failed = 9.06%, Incomplete = 90.94% | 0.00% | Failed = 0%, Incomplete = 100% | 0.00% |
| plansformer-bw-hn[5000] | Failed = 2.74%, Incomplete = 96.08% | 1.18% | Failed = 39.52%, Incomplete = 0.22% | 60.25% | Failed = 0%, Incomplete = 100% | 0.00% | Failed = 0%, Incomplete = 100% | 0.00% |
| plansformer-bw-hn[10000] | Failed = 1.22%, Incomplete = 98.78% | 0.00% | Failed = 11.66%, Incomplete = 0.06% | 88.28% | Failed = 0%, Incomplete = 100% | 0.00% | Failed = 0%, Incomplete = 100% | 0.00% |
| plansformer-bw-hn[14400] | Failed = 1.55%, Incomplete = 98.44% | 0.00% | Failed = 2.94%, Incomplete = 0% | 97.05% | Failed = 1.44%, Incomplete = 98.56% | 0.00% | Failed = 100%, Incomplete = 0% | 0.00% |
| plansformer-hn-bw[500] | Failed = 88.14%, Incomplete = 0.11% | 11.75% | Failed = 2.34%, Incomplete = 97.61% | 0.00% | Failed = 100%, Incomplete = 0% | 0.00% | Failed = 0%, Incomplete = 100% | 0.00% |
| plansformer-hn-bw[1000] | Failed = 78.97%, Incomplete = 0.42% | 20.61% | Failed = 0.28%, Incomplete = 99.72% | 0.00% | Failed = 100%, Incomplete = 0% | 0.00% | Failed = 0%, Incomplete = 100% | 0.00% |
| plansformer-hn-bw[1500] | Failed = 70.36%, Incomplete = 2.06% | 27.58% | Failed = 0.14%, Incomplete = 99.86% | 0.00% | Failed = 100%, Incomplete = 0% | 0.00% | Failed = 0%, Incomplete = 100% | 0.00% |
| plansformer-hn-bw[2000] | Failed = 53.31%, Incomplete = 0.17% | 46.53% | Failed = 0.56%, Incomplete = 99.44% | 0.00% | Failed = 100%, Incomplete = 0% | 0.00% | Failed = 0%, Incomplete = 100% | 0.00% |

| Model | | | | | | | |
|---|---|---|---|---|---|---|---|
| plansformer-hn-bw[5000] | Failed = 29.56%, Incomplete = 0.81% | 69.64% | Failed = 0.22%, Incomplete = 99.78% | 0.00% | Failed = 100%, Incomplete = 0% | 0.00% | Failed = 0%, Incomplete = 100% | 0.00% |
| plansformer-hn-bw[10000] | Failed = 9.58%, Incomplete = 0.06% | 90.36% | Failed = 4.06%, Incomplete = 95.94% | 0.00% | Failed = 100%, Incomplete = 0% | 0.00% | Failed = 0%, Incomplete = 100% | 0.00% |
| plansformer-hn-bw[14400] | Failed = 4.56%, Incomplete = 0% | 95.44% | Failed = 2.28%, Incomplete = 97.72% | 0.00% | Failed = 100%, Incomplete = 0% | 0.00% | Failed = 0%, Incomplete = 100% | 0.00% |
| plansformer-hn-gr[500] | Failed = 1.56%, Incomplete = 98.44% | 0.00% | Failed = 75.81%, Incomplete = 17.28% | 6.92% | Failed = 95.58%, Incomplete = 1.25% | 3.17% | Failed = 1.56%, Incomplete = 98.44% | 0.00% |
| plansformer-hn-gr[1000] | Failed = 1.06%, Incomplete = 98.94% | 0.00% | Failed = 31.42%, Incomplete = 55.69% | 12.89% | Failed = 89.22%, Incomplete = 0.14% | 10.64% | Failed = 0%, Incomplete = 100% | 0.00% |
| plansformer-hn-gr[1500] | Failed = 0.5%, Incomplete = 99.5% | 0.00% | Failed = 10.31%, Incomplete = 81.61% | 8.08% | Failed = 84.36%, Incomplete = 0.03% | 15.61% | Failed = 100%, Incomplete = 0.0% | 0.00% |
| plansformer-hn-gr[2000] | Failed = 1.97%, Incomplete = 98.03% | 0.00% | Failed = 1.5%, Incomplete = 91.78% | 6.72% | Failed = 77.36%, Incomplete = 0.28% | 22.36% | Failed = 100%, Incomplete = 0.0% | 0.00% |
| plansformer-hn-gr[5000] | Failed = 27.69%, Incomplete = 72.31% | 0.00% | Failed = 1.08%, Incomplete = 96.28% | 2.64% | Failed = 50.33%, Incomplete = 0.36% | 49.31% | Failed = 100%, Incomplete = 0.0% | 0.00% |
| plansformer-hn-gr[10000] | Failed = 41.14%, Incomplete = 58.86% | 0.00% | Failed = 0.0%, Incomplete = 100.0% | 0.00% | Failed = 24.25%, Incomplete = 0.17% | 75.58% | Failed = 100%, Incomplete = 0.0% | 0.00% |
| plansformer-hn-gr[14400] | Failed = 5.28%, Incomplete = 94.72% | 0.00% | Failed = 0.11%, Incomplete = 99.89% | 0.00% | Failed = 34.31%, Incomplete = 0.19% | 65.50% | Failed = 100%, Incomplete = 0.0% | 0.00% |
| plansformer-hn-dl[500] | Failed = 0.97%, Incomplete = 99.03% | 0.00% | Failed = 87.44%, Incomplete = 3.36% | 9.19% | Failed = 100.0%, Incomplete = 0.0% | 0.00% | Failed = 99.44%, Incomplete = 0.44% | 0.11% | 0.08% |
| plansformer-hn-dl[1000] | Failed = 1.11%, Incomplete = 98.89% | 0.00% | Failed = 84.86%, Incomplete = 3.89% | 11.25% | Failed = 100.0%, Incomplete = 0.0% | 0.00% | Failed = 92.47%, Incomplete = 0.56% | 6.97% | 4.64% |
| plansformer-hn-dl[1500] | Failed = 1.67%, Incomplete = 98.33% | 0.00% | Failed = 56.89%, Incomplete = 32.92% | 10.19% | Failed = 100.0%, Incomplete = 0.0% | 0.00% | Failed = 72.72%, Incomplete = 0.17% | 27.11% | 18.19% |
| plansformer-hn-dl[2000] | Failed = 1.92%, Incomplete = 98.08% | 0.00% | Failed = 50.0%, Incomplete = 38.44% | 11.56% | Failed = 100.0%, Incomplete = 0.0% | 0.00% | Failed = 59.14%, Incomplete = 0.17% | 40.69% | 30.06% |
| plansformer-hn-dl[5000] | Failed = 37.17%, Incomplete = 62.83% | 0.00% | Failed = 0.14%, Incomplete = 99.11% | 0.75% | Failed = 100.0%, Incomplete = 0.0% | 0.00% | Failed = 26.72%, Incomplete = 0.53% | 72.75% | 55.83% |

|  |  |  |  |  |  |  |  |  |  |  |
|---|---|---|---|---|---|---|---|---|---|---|
| plansformer-hn-dl[10000] | 0.00% | Failed = 0.03%, Incomplete = 99.97% | 0.00% | Failed = 0.11%, Incomplete = 99.86% | 0.03% | Failed = 100.0%, Incomplete = 0.0% | 0.00% | Failed = 13.92%, Incomplete = 0.08% | 86.00% | 67.89% |
| plansformer-hn-dl[14400] | 0.00% | Failed = 0.31%, Incomplete = 99.69% | 0.00% | Failed = 0.92%, Incomplete = 99.08% | 0.00% | Failed = 100.0%, Incomplete = 0.0% | 0.00% | Failed = 9.64%, Incomplete = 0.31% | 90.06% | 72.61% |
| plansformer-gr-bw[500] | 22.33% | Failed = 77.61%, Incomplete = 0.06% | 17.69% | Failed = 0.0%, Incomplete = 100.0% | 0.00% | Failed = 97.86%, Incomplete = 0.22% | 1.92% | Failed = 100.0%, Incomplete = 0.0% | 0.00% | 0.00% |
| plansformer-gr-bw[1000] | 48.11% | Failed = 51.89%, Incomplete = 0.0% | 35.75% | Failed = 0.03%, Incomplete = 99.97% | 0.00% | Failed = 98.06%, Incomplete = 0.44% | 1.50% | Failed = 100.0%, Incomplete = 0.0% | 0.75% | 0.00% |
| plansformer-gr-bw[1500] | 75.19% | Failed = 23.06%, Incomplete = 1.75% | 60.56% | Failed = 0.08%, Incomplete = 99.92% | 0.00% | Failed = 96.25%, Incomplete = 0.36% | 3.39% | Failed = 100.0%, Incomplete = 0.0% | 0.75% | 0.00% |
| plansformer-gr-bw[2000] | 58.33% | Failed = 40.53%, Incomplete = 1.14% | 51.89% | Failed = 0.14%, Incomplete = 99.86% | 0.00% | Failed = 98.36%, Incomplete = 0.83% | 0.81% | Failed = 100.0%, Incomplete = 0.0% | 2.56% | 0.00% |
| plansformer-gr-bw[5000] | 79.00% | Failed = 20.89%, Incomplete = 0.11% | 69.00% | Failed = 0.03%, Incomplete = 99.97% | 0.00% | Failed = 100.0%, Incomplete = 0.0% | 0.00% | Failed = 100.0%, Incomplete = 0.0% | 0.42% | 0.00% |
| plansformer-gr-bw[10000] | 93.89% | Failed = 6.11%, Incomplete = 0.0% | 91.39% | Failed = 0.0%, Incomplete = 100.0% | 0.00% | Failed = 100.0%, Incomplete = 0.0% | 0.00% | Failed = 100.0%, Incomplete = 0.0% | 0.00% | 0.00% |
| plansformer-gr-bw[14400] | 95.94% | Failed = 4.06%, Incomplete = 0.0% | 93.72% | Failed = 0.0%, Incomplete = 100.0% | 0.00% | Failed = 100.0%, Incomplete = 0.0% | 0.00% | Failed = 100.0%, Incomplete = 0.0% | 0.00% | 0.00% |
| plansformer-gr-hn[500] | 0.00% | Failed = 0.0%, Incomplete = 100.0% | 0.00% | Failed = 96.33%, Incomplete = 0.06% | 3.61% | Failed = 96.03%, Incomplete = 0.47% | 3.50% | Failed = 100.0%, Incomplete = 0.0% | 1.39% | 0.00% |
| plansformer-gr-hn[1000] | 0.00% | Failed = 0.0%, Incomplete = 100.0% | 0.00% | Failed = 90.64%, Incomplete = 0.03% | 9.33% | Failed = 96.06%, Incomplete = 0.58% | 3.36% | Failed = 100.0%, Incomplete = 0.0% | 2.06% | 0.00% |
| plansformer-gr-hn[1500] | 0.00% | Failed = 0.0%, Incomplete = 100.0% | 0.00% | Failed = 75.19%, Incomplete = 0.03% | 24.78% | Failed = 96.22%, Incomplete = 0.86% | 2.92% | Failed = 100.0%, Incomplete = 0.0% | 1.03% | 0.00% |
| plansformer-gr-hn[2000] | 0.00% | Failed = 0.03%, Incomplete = 99.97% | 0.00% | Failed = 65.36%, Incomplete = 0.0% | 34.64% | Failed = 97.58%, Incomplete = 0.83% | 1.58% | Failed = 100.0%, Incomplete = 0.0% | 1.06% | 0.00% |
| plansformer-gr-hn[5000] | 0.00% | Failed = 0.03%, Incomplete = 99.97% | 0.00% | Failed = 28.56%, Incomplete = 0.08% | 71.36% | Failed = 97.89%, Incomplete = 2.11% | 0.00% | Failed = 100.0%, Incomplete = 0.0% | 0.00% | 0.00% |
| plansformer-gr-hn[10000] | 0.00% | Failed = 0.75%, Incomplete = 99.25% | 0.00% | Failed = 11.53%, Incomplete = 0.0% | 88.47% | Failed = 63.72%, Incomplete = 36.28% | 0.00% | Failed = 100.0%, Incomplete = 0.0% | 0.00% | 0.00% |

| | | | | | | | |
|---|---|---|---|---|---|---|---|
| plansformer-gr-hn[14400] | Failed = 0.08%, Incomplete = 99.92% | 0.00% | Failed = 8.94%, Incomplete = 0.0% | 89.86% | Failed = 6.0%, Incomplete = 94.0% | 0.00% | Failed = 100.0%, Incomplete = 0.0% | 0.00% |
| plansformer-gr-dl[500] | Failed = 0.06%, Incomplete = 99.94% | 0.00% | Failed = 0.11%, Incomplete = 99.89% | 0.00% | Failed = 25.28%, Incomplete = 0.0% | 74.72% | Failed = 64.28%, Incomplete = 4.64% | 31.08% | 18.44% |
| plansformer-gr-dl[1000] | Failed = 0.39%, Incomplete = 99.61% | 0.00% | Failed = 0.0%, Incomplete = 100.0% | 0.00% | Failed = 25.22%, Incomplete = 0.03% | 74.75% | Failed = 48.92%, Incomplete = 0.28% | 50.81% | 37.72% |
| plansformer-gr-dl[1500] | Failed = 0.06%, Incomplete = 99.94% | 0.00% | Failed = 0.03%, Incomplete = 99.97% | 0.00% | Failed = 33.67%, Incomplete = 0.03% | 66.31% | Failed = 38.47%, Incomplete = 0.11% | 61.42% | 40.19% |
| plansformer-gr-dl[2000] | Failed = 0.06%, Incomplete = 99.94% | 0.00% | Failed = 0.06%, Incomplete = 99.94% | 0.00% | Failed = 43.14%, Incomplete = 3.17% | 53.69% | Failed = 34.64%, Incomplete = 0.33% | 65.03% | 45.86% |
| plansformer-gr-dl[5000] | Failed = 0.08%, Incomplete = 99.92% | 0.00% | Failed = 0.03%, Incomplete = 99.97% | 0.00% | Failed = 96.0%, Incomplete = 1.42% | 2.58% | Failed = 11.08%, Incomplete = 12.56% | 76.36% | 60.72% |
| plansformer-gr-dl[10000] | Failed = 1.03%, Incomplete = 98.97% | 0.00% | Failed = 1.03%, Incomplete = 98.97% | 0.00% | Failed = 99.94%, Incomplete = 0.06% | 0.00% | Failed = 9.31%, Incomplete = 0.25% | 90.44% | 74.06% |
| plansformer-gr-dl[14400] | Failed = 0.86%, Incomplete = 99.14% | 0.00% | Failed = 3.22%, Incomplete = 96.78% | 0.00% | Failed = 98.5%, Incomplete = 1.5% | 0.00% | Failed = 7.94%, Incomplete = 0.08% | 91.97% | 76.14% |
| plansformer-dl-bw[500] | Failed = 79.17%, Incomplete = 0.0% | 15.89% | Failed = 0.0%, Incomplete = 100.0% | 0.00% | Failed = 67.89%, Incomplete = 32.11% | 0.00% | Failed = 100.0%, Incomplete = 0.0% | 0.00% | 0.00% |
| plansformer-dl-bw[1000] | Failed = 59.36%, Incomplete = 0.28% | 40.36% | Failed = 0.0%, Incomplete = 100.0% | 0.00% | Failed = 98.36%, Incomplete = 1.64% | 0.00% | Failed = 99.53%, Incomplete = 0.44% | 0.03% | 0.00% |
| plansformer-dl-bw[1500] | Failed = 60.08%, Incomplete = 0.06% | 39.86% | Failed = 0.0%, Incomplete = 100.0% | 32.47% | Failed = 100.0%, Incomplete = 0.0% | 0.00% | Failed = 100.0%, Incomplete = 0.0% | 0.00% | 0.00% |
| plansformer-dl-bw[2000] | Failed = 31.22%, Incomplete = 0.78% | 68.00% | Failed = 0.0%, Incomplete = 100.0% | 62.42% | Failed = 100.0%, Incomplete = 0.0% | 0.00% | Failed = 100.0%, Incomplete = 0.0% | 0.00% | 0.00% |
| plansformer-dl-bw[5000] | Failed = 9.03%, Incomplete = 0.0% | 90.97% | Failed = 0.47%, Incomplete = 99.53% | 82.69% | Failed = 100.0%, Incomplete = 0.0% | 0.00% | Failed = 100.0%, Incomplete = 0.0% | 0.00% | 0.00% |
| plansformer-dl-bw[10000] | Failed = 9.81%, Incomplete = 0.03% | 90.17% | Failed = 0.06%, Incomplete = 99.94% | 88.22% | Failed = 100.0%, Incomplete = 0.0% | 0.00% | Failed = 100.0%, Incomplete = 0.0% | 0.00% | 0.00% |
| plansformer-dl-bw[14400] | Failed = 6.81%, Incomplete = 0.0% | 93.19% | Failed = 0.0%, Incomplete = 100.0% | 91.28% | Failed = 100.0%, Incomplete = 0.0% | 0.00% | Failed = 100.0%, Incomplete = 0.0% | 0.00% | 0.00% |

| Model | | | | | | | |
|---|---|---|---|---|---|---|---|
| plansformer-dl-hn[500] | 0.00% | Failed = 1.31%, Incomplete = 98.69% | 0.00% | Failed = 98.94%, Incomplete = 0.06% | 0.61% | Failed = 99.89%, Incomplete = 0.11% | 0.00% | 33.25% | Failed = 48.72%, Incomplete = 18.03% | 28.97% |
| plansformer-dl-hn[1000] | 0.00% | Failed = 1.69%, Incomplete = 98.31% | 0.00% | Failed = 96.69%, Incomplete = 0.0% | 3.28% | Failed = 3.83%, Incomplete = 96.17% | 0.00% | 10.28% | Failed = 49.78%, Incomplete = 39.94% | 9.39% |
| plansformer-dl-hn[1500] | 0.00% | Failed = 0.0%, Incomplete = 100.0% | 0.00% | Failed = 83.61%, Incomplete = 0.11% | 15.69% | Failed = 0.03%, Incomplete = 99.97% | 0.00% | 11.72% | Failed = 41.44%, Incomplete = 46.83% | 11.03% |
| plansformer-dl-hn[2000] | 0.00% | Failed = 0.17%, Incomplete = 99.83% | 0.00% | Failed = 86.69%, Incomplete = 0.28% | 12.83% | Failed = 4.97%, Incomplete = 95.03% | 0.00% | 5.50% | Failed = 54.81%, Incomplete = 39.69% | 5.17% |
| plansformer-dl-hn[5000] | 0.00% | Failed = 2.19%, Incomplete = 97.81% | 0.00% | Failed = 56.97%, Incomplete = 0.06% | 38.75% | Failed = 2.69%, Incomplete = 97.31% | 0.00% | 0.00% | Failed = 99.22%, Incomplete = 0.78% | 0.00% |
| plansformer-dl-hn[10000] | 0.00% | Failed = 1.28%, Incomplete = 98.72% | 0.00% | Failed = 10.31%, Incomplete = 0.03% | 87.72% | Failed = 0.22%, Incomplete = 99.78% | 0.00% | 0.00% | Failed = 100.0%, Incomplete = 0.0% | 0.00% |
| plansformer-dl-hn[14400] | 0.00% | Failed = 0.53%, Incomplete = 99.47% | 0.00% | Failed = 18.03%, Incomplete = 0.11% | 79.81% | Failed = 1.06%, Incomplete = 98.94% | 0.00% | 0.00% | Failed = 100.0%, Incomplete = 0.0% | 0.00% |
| plansformer-dl-gr[500] | 0.00% | Failed = 3.5%, Incomplete = 96.5% | 0.00% | Failed = 0.0%, Incomplete = 100.0% | 0.00% | Failed = 81.36%, Incomplete = 0.25% | 18.39% | 29.11% | Failed = 39.14%, Incomplete = 31.75% | 24.89% |
| plansformer-dl-gr[1000] | 0.00% | Failed = 0.0%, Incomplete = 100.0% | 0.00% | Failed = 0.0%, Incomplete = 100.0% | 0.00% | Failed = 51.22%, Incomplete = 0.06% | 48.72% | 45.92% | Failed = 45.03%, Incomplete = 9.06% | 38.14% |
| plansformer-dl-gr[1500] | 0.00% | Failed = 0.22%, Incomplete = 99.78% | 0.00% | Failed = 0.0%, Incomplete = 100.0% | 0.00% | Failed = 37.56%, Incomplete = 0.97% | 61.47% | 20.22% | Failed = 68.03%, Incomplete = 11.75% | 15.97% |
| plansformer-dl-gr[2000] | 0.00% | Failed = 1.72%, Incomplete = 98.28% | 0.00% | Failed = 0.0%, Incomplete = 100.0% | 0.00% | Failed = 37.83%, Incomplete = 0.08% | 62.08% | 4.50% | Failed = 91.81%, Incomplete = 3.69% | 3.67% |
| plansformer-dl-gr[5000] | 0.00% | Failed = 0.0%, Incomplete = 100.0% | 0.00% | Failed = 0.0%, Incomplete = 100.0% | 0.00% | Failed = 22.25%, Incomplete = 0.03% | 77.72% | 0.00% | Failed = 99.97%, Incomplete = 0.03% | 0.00% |
| plansformer-dl-gr[10000] | 0.00% | Failed = 0.0%, Incomplete = 100.0% | 0.00% | Failed = 0.0%, Incomplete = 100.0% | 0.00% | Failed = 15.0%, Incomplete = 0.0% | 85.00% | 0.00% | Failed = 100.0%, Incomplete = 0.0% | 0.00% |
| plansformer-dl-gr[14400] | 0.00% | Failed = 0.06%, Incomplete = 99.94% | 0.00% | Failed = 0.0%, Incomplete = 100.0% | 0.00% | Failed = 14.5%, Incomplete = 0.03% | 85.47% | 0.00% | Failed = 100.0%, Incomplete = 0.0% | 0.00% |

Further fine-tuned models

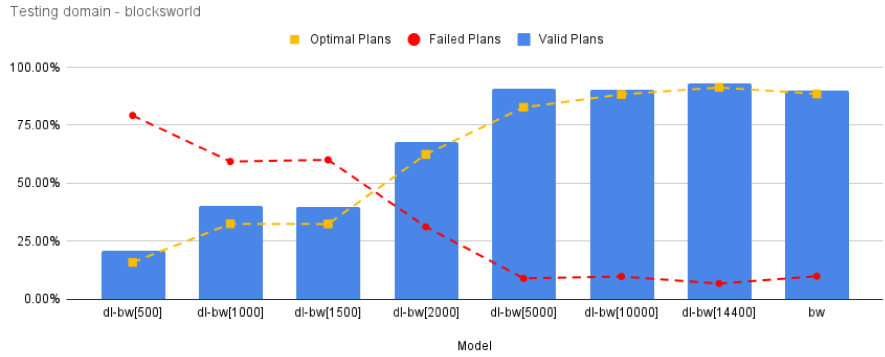

Figure 7: Plansformer-dl variants performance on blocksworld at various stages of fine-tuning

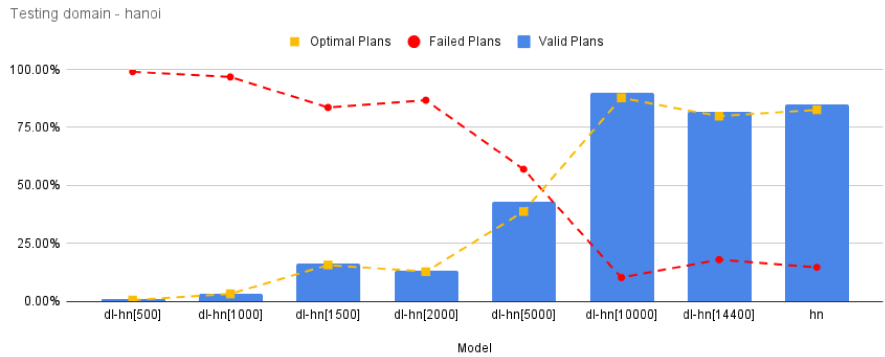

Figure 8: Plansformer-dl variants performance on hanoi at various stages of fine-tuning

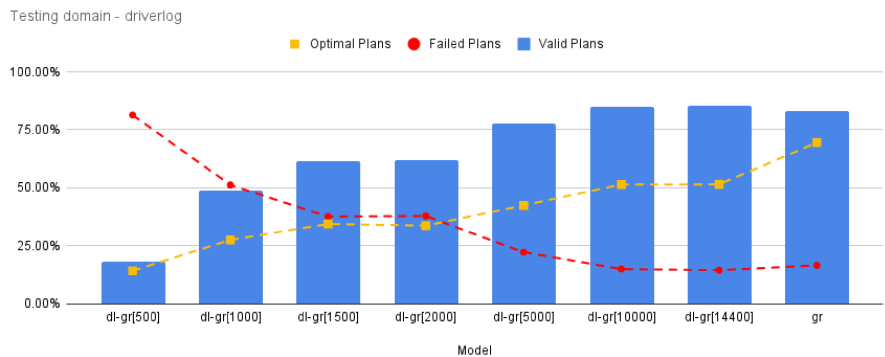

Figure 9: Plansformer-dl variants performance on grippers at various stages of fine-tuning

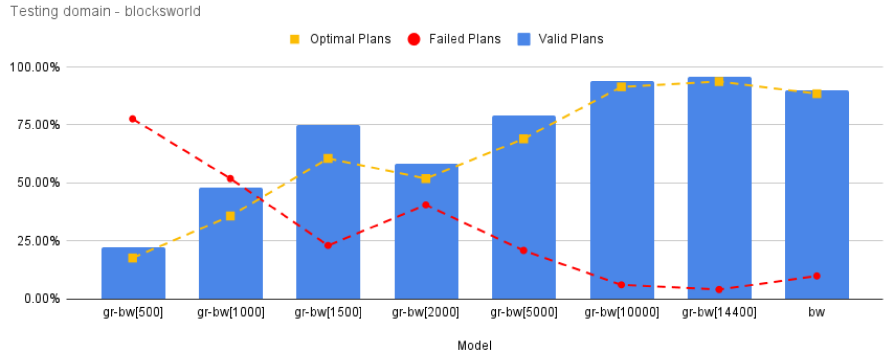

Figure 10: Plansformer-gr variants performance on blocksworld at various stages of fine-tuning

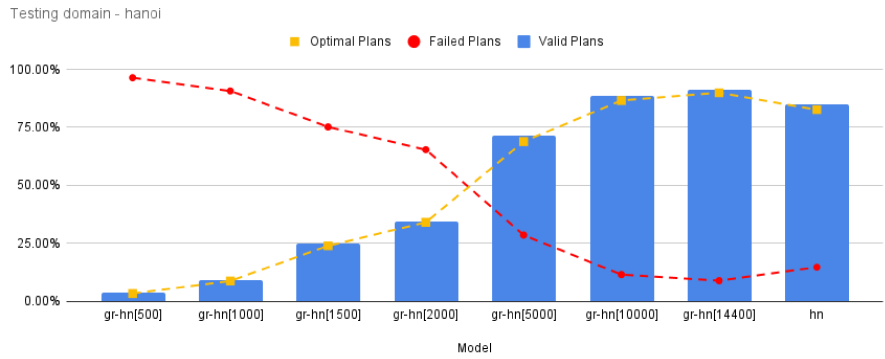

Figure 11: Plansformer-gr variants performance on hanoi at various stages of fine-tuning

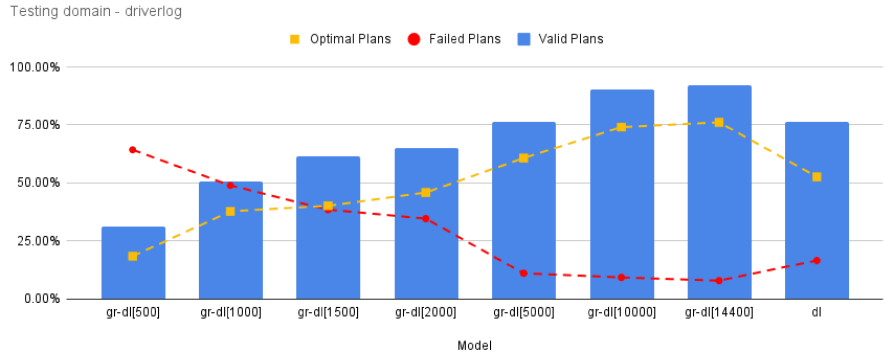

Figure 12: Plansformer-gr variants performance on driverlog at various stages of fine-tuning

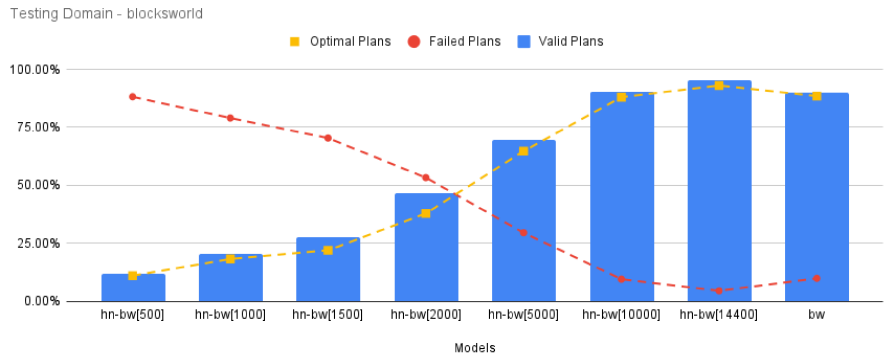

Figure 13: Plansformer-hn variants performance on blocksworld at various stages of fine-tuning

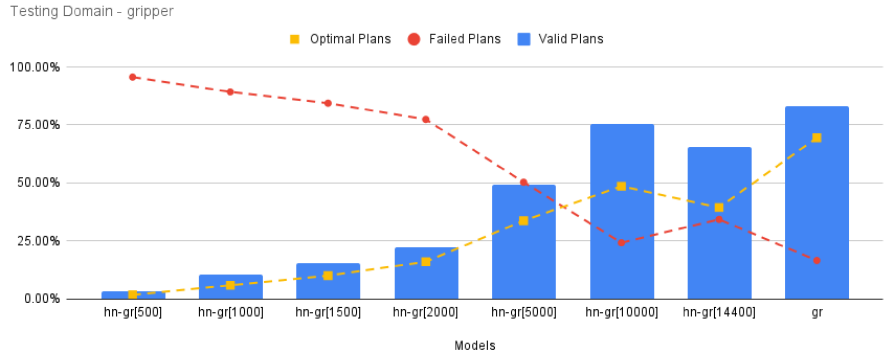

Figure 14: Plansformer-hn variants performance on gripper at various stages of fine-tuning

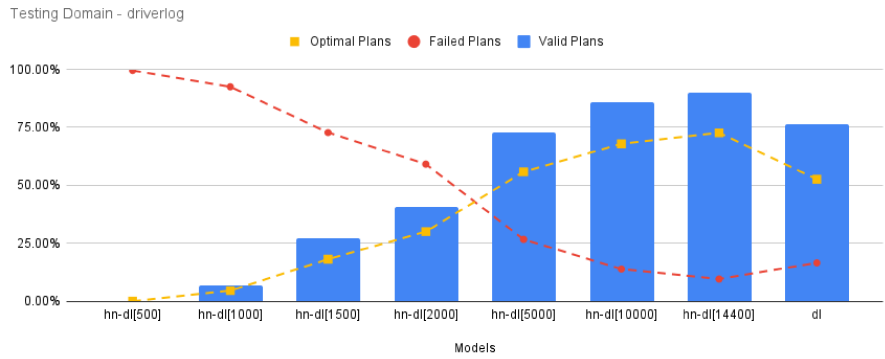

Figure 15: Plansformer-dl variants performance on driverlog at various stages of fine-tuning

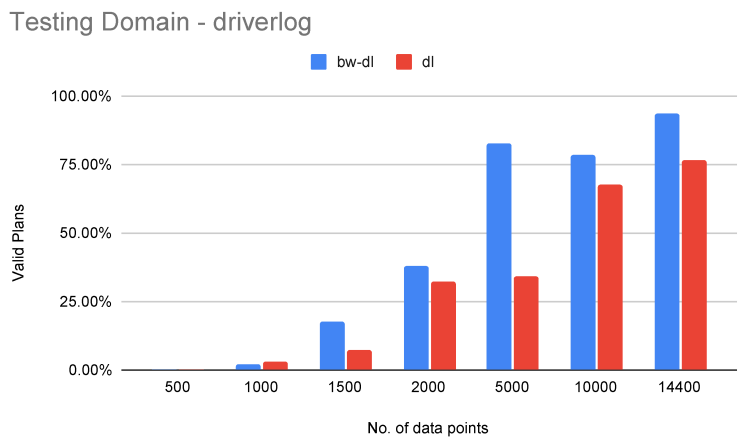

Figure 16: Comparison of valid plans generated by Plansformer-bw-dl derived models with Plansformer-dl trained using similar data points.

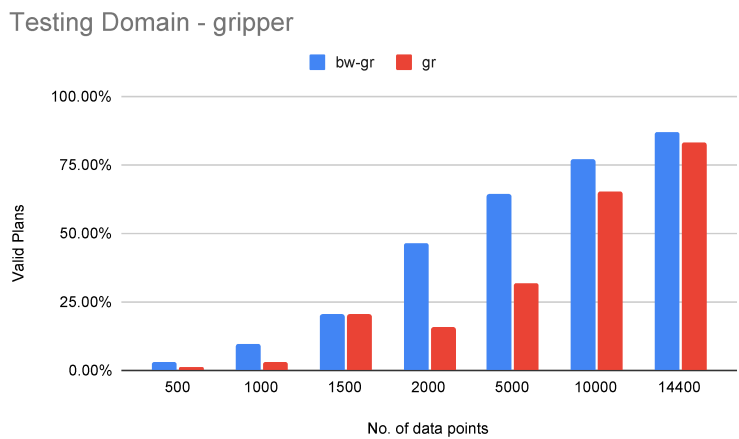

Figure 17: Comparison of valid plans generated by Plansformer-bw-gr derived models with Plansformer-gr trained using similar data points.

```
T5ForConditionalGeneration(
  (shared): Embedding(32100, 768)
  (encoder): T5Stack(
    (embed_tokens): Embedding(32100, 768)
    (block): ModuleList(
      (0): T5Block(
        (layer): ModuleList(
          (0): T5LayerSelfAttention(
            (SelfAttention): T5Attention(
              (q): Linear(in_features=768, out_features=768, bias=False)
              (k): Linear(in_features=768, out_features=768, bias=False)
              (v): Linear(in_features=768, out_features=768, bias=False)
              (o): Linear(in_features=768, out_features=768, bias=False)
              (relative_attention_bias): Embedding(32, 12)
            )
            (layer_norm): T5LayerNorm()
            (dropout): Dropout(p=0.1, inplace=False)
          )
          (1): T5LayerFF(
            (DenseReluDense): T5DenseActDense(
              (wi): Linear(in_features=768, out_features=3072, bias=False)
              (wo): Linear(in_features=3072, out_features=768, bias=False)
              (dropout): Dropout(p=0.1, inplace=False)
              (act): ReLU()
            )
            (layer_norm): T5LayerNorm()
            (dropout): Dropout(p=0.1, inplace=False)
          )
        )
      )
      (1): T5Block(
        (layer): ModuleList(
          (0): T5LayerSelfAttention(
            (SelfAttention): T5Attention(
              (q): Linear(in_features=768, out_features=768, bias=False)
              (k): Linear(in_features=768, out_features=768, bias=False)
              (v): Linear(in_features=768, out_features=768, bias=False)
              (o): Linear(in_features=768, out_features=768, bias=False)
            )
            (layer_norm): T5LayerNorm()
            (dropout): Dropout(p=0.1, inplace=False)
          )
          (1): T5LayerFF(
            (DenseReluDense): T5DenseActDense(
              (wi): Linear(in_features=768, out_features=3072, bias=False)
              (wo): Linear(in_features=3072, out_features=768, bias=False)
              (dropout): Dropout(p=0.1, inplace=False)
              (act): ReLU()
            )
            (layer_norm): T5LayerNorm()
            (dropout): Dropout(p=0.1, inplace=False)
          )
        )
      )
      (2): T5Block(
        (layer): ModuleList(
          (0): T5LayerSelfAttention(
            (SelfAttention): T5Attention(
              (q): Linear(in_features=768, out_features=768, bias=False)
              (k): Linear(in_features=768, out_features=768, bias=False)
              (v): Linear(in_features=768, out_features=768, bias=False)
              (o): Linear(in_features=768, out_features=768, bias=False)
```

Figure 18: Plansformer Layers

```
          )
          (layer_norm): T5LayerNorm()
          (dropout): Dropout(p=0.1, inplace=False)
        )
        (1): T5LayerFF(
          (DenseReluDense): T5DenseActDense(
            (wi): Linear(in_features=768, out_features=3072, bias=False)
            (wo): Linear(in_features=3072, out_features=768, bias=False)
            (dropout): Dropout(p=0.1, inplace=False)
            (act): ReLU()
          )
          (layer_norm): T5LayerNorm()
          (dropout): Dropout(p=0.1, inplace=False)
        )
      )
    )
    (3): T5Block(
      (layer): ModuleList(
        (0): T5LayerSelfAttention(
          (SelfAttention): T5Attention(
            (q): Linear(in_features=768, out_features=768, bias=False)
            (k): Linear(in_features=768, out_features=768, bias=False)
            (v): Linear(in_features=768, out_features=768, bias=False)
            (o): Linear(in_features=768, out_features=768, bias=False)
          )
          (layer_norm): T5LayerNorm()
          (dropout): Dropout(p=0.1, inplace=False)
        )
        (1): T5LayerFF(
          (DenseReluDense): T5DenseActDense(
            (wi): Linear(in_features=768, out_features=3072, bias=False)
            (wo): Linear(in_features=3072, out_features=768, bias=False)
            (dropout): Dropout(p=0.1, inplace=False)
            (act): ReLU()
          )
          (layer_norm): T5LayerNorm()
          (dropout): Dropout(p=0.1, inplace=False)
        )
      )
    )
    (4): T5Block(
      (layer): ModuleList(
        (0): T5LayerSelfAttention(
          (SelfAttention): T5Attention(
            (q): Linear(in_features=768, out_features=768, bias=False)
            (k): Linear(in_features=768, out_features=768, bias=False)
            (v): Linear(in_features=768, out_features=768, bias=False)
            (o): Linear(in_features=768, out_features=768, bias=False)
          )
          (layer_norm): T5LayerNorm()
          (dropout): Dropout(p=0.1, inplace=False)
        )
        (1): T5LayerFF(
          (DenseReluDense): T5DenseActDense(
            (wi): Linear(in_features=768, out_features=3072, bias=False)
            (wo): Linear(in_features=3072, out_features=768, bias=False)
            (dropout): Dropout(p=0.1, inplace=False)
            (act): ReLU()
          )
          (layer_norm): T5LayerNorm()
          (dropout): Dropout(p=0.1, inplace=False)
        )
```

```
          )
        )
        (5): T5Block(
          (layer): ModuleList(
            (0): T5LayerSelfAttention(
              (SelfAttention): T5Attention(
                (q): Linear(in_features=768, out_features=768, bias=False)
                (k): Linear(in_features=768, out_features=768, bias=False)
                (v): Linear(in_features=768, out_features=768, bias=False)
                (o): Linear(in_features=768, out_features=768, bias=False)
              )
              (layer_norm): T5LayerNorm()
              (dropout): Dropout(p=0.1, inplace=False)
            )
            (1): T5LayerFF(
              (DenseReluDense): T5DenseActDense(
                (wi): Linear(in_features=768, out_features=3072, bias=False)
                (wo): Linear(in_features=3072, out_features=768, bias=False)
                (dropout): Dropout(p=0.1, inplace=False)
                (act): ReLU()
              )
              (layer_norm): T5LayerNorm()
              (dropout): Dropout(p=0.1, inplace=False)
            )
          )
        )
        (6): T5Block(
          (layer): ModuleList(
            (0): T5LayerSelfAttention(
              (SelfAttention): T5Attention(
                (q): Linear(in_features=768, out_features=768, bias=False)
                (k): Linear(in_features=768, out_features=768, bias=False)
                (v): Linear(in_features=768, out_features=768, bias=False)
                (o): Linear(in_features=768, out_features=768, bias=False)
              )
              (layer_norm): T5LayerNorm()
              (dropout): Dropout(p=0.1, inplace=False)
            )
            (1): T5LayerFF(
              (DenseReluDense): T5DenseActDense(
                (wi): Linear(in_features=768, out_features=3072, bias=False)
                (wo): Linear(in_features=3072, out_features=768, bias=False)
                (dropout): Dropout(p=0.1, inplace=False)
                (act): ReLU()
              )
              (layer_norm): T5LayerNorm()
              (dropout): Dropout(p=0.1, inplace=False)
            )
          )
        )
        (7): T5Block(
          (layer): ModuleList(
            (0): T5LayerSelfAttention(
              (SelfAttention): T5Attention(
                (q): Linear(in_features=768, out_features=768, bias=False)
                (k): Linear(in_features=768, out_features=768, bias=False)
                (v): Linear(in_features=768, out_features=768, bias=False)
                (o): Linear(in_features=768, out_features=768, bias=False)
              )
              (layer_norm): T5LayerNorm()
              (dropout): Dropout(p=0.1, inplace=False)
            )
```

```
      (1): T5LayerFF(
        (DenseReluDense): T5DenseActDense(
          (wi): Linear(in_features=768, out_features=3072, bias=False)
          (wo): Linear(in_features=3072, out_features=768, bias=False)
          (dropout): Dropout(p=0.1, inplace=False)
          (act): ReLU()
        )
        (layer_norm): T5LayerNorm()
        (dropout): Dropout(p=0.1, inplace=False)
      )
    )
  )
  (8): T5Block(
    (layer): ModuleList(
      (0): T5LayerSelfAttention(
        (SelfAttention): T5Attention(
          (q): Linear(in_features=768, out_features=768, bias=False)
          (k): Linear(in_features=768, out_features=768, bias=False)
          (v): Linear(in_features=768, out_features=768, bias=False)
          (o): Linear(in_features=768, out_features=768, bias=False)
        )
        (layer_norm): T5LayerNorm()
        (dropout): Dropout(p=0.1, inplace=False)
      )
      (1): T5LayerFF(
        (DenseReluDense): T5DenseActDense(
          (wi): Linear(in_features=768, out_features=3072, bias=False)
          (wo): Linear(in_features=3072, out_features=768, bias=False)
          (dropout): Dropout(p=0.1, inplace=False)
          (act): ReLU()
        )
        (layer_norm): T5LayerNorm()
        (dropout): Dropout(p=0.1, inplace=False)
      )
    )
  )
  (9): T5Block(
    (layer): ModuleList(
      (0): T5LayerSelfAttention(
        (SelfAttention): T5Attention(
          (q): Linear(in_features=768, out_features=768, bias=False)
          (k): Linear(in_features=768, out_features=768, bias=False)
          (v): Linear(in_features=768, out_features=768, bias=False)
          (o): Linear(in_features=768, out_features=768, bias=False)
        )
        (layer_norm): T5LayerNorm()
        (dropout): Dropout(p=0.1, inplace=False)
      )
      (1): T5LayerFF(
        (DenseReluDense): T5DenseActDense(
          (wi): Linear(in_features=768, out_features=3072, bias=False)
          (wo): Linear(in_features=3072, out_features=768, bias=False)
          (dropout): Dropout(p=0.1, inplace=False)
          (act): ReLU()
        )
        (layer_norm): T5LayerNorm()
        (dropout): Dropout(p=0.1, inplace=False)
      )
    )
  )
  (10): T5Block(
    (layer): ModuleList(
```

```
          (0): T5LayerSelfAttention(
            (SelfAttention): T5Attention(
              (q): Linear(in_features=768, out_features=768, bias=False)
              (k): Linear(in_features=768, out_features=768, bias=False)
              (v): Linear(in_features=768, out_features=768, bias=False)
              (o): Linear(in_features=768, out_features=768, bias=False)
            )
            (layer_norm): T5LayerNorm()
            (dropout): Dropout(p=0.1, inplace=False)
          )
          (1): T5LayerFF(
            (DenseReluDense): T5DenseActDense(
              (wi): Linear(in_features=768, out_features=3072, bias=False)
              (wo): Linear(in_features=3072, out_features=768, bias=False)
              (dropout): Dropout(p=0.1, inplace=False)
              (act): ReLU()
            )
            (layer_norm): T5LayerNorm()
            (dropout): Dropout(p=0.1, inplace=False)
          )
        )
      )
      (11): T5Block(
        (layer): ModuleList(
          (0): T5LayerSelfAttention(
            (SelfAttention): T5Attention(
              (q): Linear(in_features=768, out_features=768, bias=False)
              (k): Linear(in_features=768, out_features=768, bias=False)
              (v): Linear(in_features=768, out_features=768, bias=False)
              (o): Linear(in_features=768, out_features=768, bias=False)
            )
            (layer_norm): T5LayerNorm()
            (dropout): Dropout(p=0.1, inplace=False)
          )
          (1): T5LayerFF(
            (DenseReluDense): T5DenseActDense(
              (wi): Linear(in_features=768, out_features=3072, bias=False)
              (wo): Linear(in_features=3072, out_features=768, bias=False)
              (dropout): Dropout(p=0.1, inplace=False)
              (act): ReLU()
            )
            (layer_norm): T5LayerNorm()
            (dropout): Dropout(p=0.1, inplace=False)
          )
        )
      )
    )
    (final_layer_norm): T5LayerNorm()
    (dropout): Dropout(p=0.1, inplace=False)
  )
  (decoder): T5Stack(
    (embed_tokens): Embedding(32100, 768)
    (block): ModuleList(
      (0): T5Block(
        (layer): ModuleList(
          (0): T5LayerSelfAttention(
            (SelfAttention): T5Attention(
              (q): Linear(in_features=768, out_features=768, bias=False)
              (k): Linear(in_features=768, out_features=768, bias=False)
              (v): Linear(in_features=768, out_features=768, bias=False)
              (o): Linear(in_features=768, out_features=768, bias=False)
              (relative_attention_bias): Embedding(32, 12)
```

```
              )
              (layer_norm): T5LayerNorm()
              (dropout): Dropout(p=0.1, inplace=False)
            )
            (1): T5LayerCrossAttention(
              (EncDecAttention): T5Attention(
                (q): Linear(in_features=768, out_features=768, bias=False)
                (k): Linear(in_features=768, out_features=768, bias=False)
                (v): Linear(in_features=768, out_features=768, bias=False)
                (o): Linear(in_features=768, out_features=768, bias=False)
              )
              (layer_norm): T5LayerNorm()
              (dropout): Dropout(p=0.1, inplace=False)
            )
            (2): T5LayerFF(
              (DenseReluDense): T5DenseActDense(
                (wi): Linear(in_features=768, out_features=3072, bias=False)
                (wo): Linear(in_features=3072, out_features=768, bias=False)
                (dropout): Dropout(p=0.1, inplace=False)
                (act): ReLU()
              )
              (layer_norm): T5LayerNorm()
              (dropout): Dropout(p=0.1, inplace=False)
            )
          )
        )
        (1): T5Block(
          (layer): ModuleList(
            (0): T5LayerSelfAttention(
              (SelfAttention): T5Attention(
                (q): Linear(in_features=768, out_features=768, bias=False)
                (k): Linear(in_features=768, out_features=768, bias=False)
                (v): Linear(in_features=768, out_features=768, bias=False)
                (o): Linear(in_features=768, out_features=768, bias=False)
              )
              (layer_norm): T5LayerNorm()
              (dropout): Dropout(p=0.1, inplace=False)
            )
            (1): T5LayerCrossAttention(
              (EncDecAttention): T5Attention(
                (q): Linear(in_features=768, out_features=768, bias=False)
                (k): Linear(in_features=768, out_features=768, bias=False)
                (v): Linear(in_features=768, out_features=768, bias=False)
                (o): Linear(in_features=768, out_features=768, bias=False)
              )
              (layer_norm): T5LayerNorm()
              (dropout): Dropout(p=0.1, inplace=False)
            )
            (2): T5LayerFF(
              (DenseReluDense): T5DenseActDense(
                (wi): Linear(in_features=768, out_features=3072, bias=False)
                (wo): Linear(in_features=3072, out_features=768, bias=False)
                (dropout): Dropout(p=0.1, inplace=False)
                (act): ReLU()
              )
              (layer_norm): T5LayerNorm()
              (dropout): Dropout(p=0.1, inplace=False)
            )
          )
        )
        (2): T5Block(
          (layer): ModuleList(
```

```
        (0): T5LayerSelfAttention(
          (SelfAttention): T5Attention(
            (q): Linear(in_features=768, out_features=768, bias=False)
            (k): Linear(in_features=768, out_features=768, bias=False)
            (v): Linear(in_features=768, out_features=768, bias=False)
            (o): Linear(in_features=768, out_features=768, bias=False)
          )
          (layer_norm): T5LayerNorm()
          (dropout): Dropout(p=0.1, inplace=False)
        )
        (1): T5LayerCrossAttention(
          (EncDecAttention): T5Attention(
            (q): Linear(in_features=768, out_features=768, bias=False)
            (k): Linear(in_features=768, out_features=768, bias=False)
            (v): Linear(in_features=768, out_features=768, bias=False)
            (o): Linear(in_features=768, out_features=768, bias=False)
          )
          (layer_norm): T5LayerNorm()
          (dropout): Dropout(p=0.1, inplace=False)
        )
        (2): T5LayerFF(
          (DenseReluDense): T5DenseActDense(
            (wi): Linear(in_features=768, out_features=3072, bias=False)
            (wo): Linear(in_features=3072, out_features=768, bias=False)
            (dropout): Dropout(p=0.1, inplace=False)
            (act): ReLU()
          )
          (layer_norm): T5LayerNorm()
          (dropout): Dropout(p=0.1, inplace=False)
        )
      )
    )
    (3): T5Block(
      (layer): ModuleList(
        (0): T5LayerSelfAttention(
          (SelfAttention): T5Attention(
            (q): Linear(in_features=768, out_features=768, bias=False)
            (k): Linear(in_features=768, out_features=768, bias=False)
            (v): Linear(in_features=768, out_features=768, bias=False)
            (o): Linear(in_features=768, out_features=768, bias=False)
          )
          (layer_norm): T5LayerNorm()
          (dropout): Dropout(p=0.1, inplace=False)
        )
        (1): T5LayerCrossAttention(
          (EncDecAttention): T5Attention(
            (q): Linear(in_features=768, out_features=768, bias=False)
            (k): Linear(in_features=768, out_features=768, bias=False)
            (v): Linear(in_features=768, out_features=768, bias=False)
            (o): Linear(in_features=768, out_features=768, bias=False)
          )
          (layer_norm): T5LayerNorm()
          (dropout): Dropout(p=0.1, inplace=False)
        )
        (2): T5LayerFF(
          (DenseReluDense): T5DenseActDense(
            (wi): Linear(in_features=768, out_features=3072, bias=False)
            (wo): Linear(in_features=3072, out_features=768, bias=False)
            (dropout): Dropout(p=0.1, inplace=False)
            (act): ReLU()
          )
          (layer_norm): T5LayerNorm()
```

```
          (dropout): Dropout(p=0.1, inplace=False)
        )
      )
    )
    (4): T5Block(
      (layer): ModuleList(
        (0): T5LayerSelfAttention(
          (SelfAttention): T5Attention(
            (q): Linear(in_features=768, out_features=768, bias=False)
            (k): Linear(in_features=768, out_features=768, bias=False)
            (v): Linear(in_features=768, out_features=768, bias=False)
            (o): Linear(in_features=768, out_features=768, bias=False)
          )
          (layer_norm): T5LayerNorm()
          (dropout): Dropout(p=0.1, inplace=False)
        )
        (1): T5LayerCrossAttention(
          (EncDecAttention): T5Attention(
            (q): Linear(in_features=768, out_features=768, bias=False)
            (k): Linear(in_features=768, out_features=768, bias=False)
            (v): Linear(in_features=768, out_features=768, bias=False)
            (o): Linear(in_features=768, out_features=768, bias=False)
          )
          (layer_norm): T5LayerNorm()
          (dropout): Dropout(p=0.1, inplace=False)
        )
        (2): T5LayerFF(
          (DenseReluDense): T5DenseActDense(
            (wi): Linear(in_features=768, out_features=3072, bias=False)
            (wo): Linear(in_features=3072, out_features=768, bias=False)
            (dropout): Dropout(p=0.1, inplace=False)
            (act): ReLU()
          )
          (layer_norm): T5LayerNorm()
          (dropout): Dropout(p=0.1, inplace=False)
        )
      )
    )
    (5): T5Block(
      (layer): ModuleList(
        (0): T5LayerSelfAttention(
          (SelfAttention): T5Attention(
            (q): Linear(in_features=768, out_features=768, bias=False)
            (k): Linear(in_features=768, out_features=768, bias=False)
            (v): Linear(in_features=768, out_features=768, bias=False)
            (o): Linear(in_features=768, out_features=768, bias=False)
          )
          (layer_norm): T5LayerNorm()
          (dropout): Dropout(p=0.1, inplace=False)
        )
        (1): T5LayerCrossAttention(
          (EncDecAttention): T5Attention(
            (q): Linear(in_features=768, out_features=768, bias=False)
            (k): Linear(in_features=768, out_features=768, bias=False)
            (v): Linear(in_features=768, out_features=768, bias=False)
            (o): Linear(in_features=768, out_features=768, bias=False)
          )
          (layer_norm): T5LayerNorm()
          (dropout): Dropout(p=0.1, inplace=False)
        )
        (2): T5LayerFF(
          (DenseReluDense): T5DenseActDense(
```

```
          (wi): Linear(in_features=768, out_features=3072, bias=False)
          (wo): Linear(in_features=3072, out_features=768, bias=False)
          (dropout): Dropout(p=0.1, inplace=False)
          (act): ReLU()
        )
        (layer_norm): T5LayerNorm()
        (dropout): Dropout(p=0.1, inplace=False)
      )
    )
  )
  (6): T5Block(
    (layer): ModuleList(
      (0): T5LayerSelfAttention(
        (SelfAttention): T5Attention(
          (q): Linear(in_features=768, out_features=768, bias=False)
          (k): Linear(in_features=768, out_features=768, bias=False)
          (v): Linear(in_features=768, out_features=768, bias=False)
          (o): Linear(in_features=768, out_features=768, bias=False)
        )
        (layer_norm): T5LayerNorm()
        (dropout): Dropout(p=0.1, inplace=False)
      )
      (1): T5LayerCrossAttention(
        (EncDecAttention): T5Attention(
          (q): Linear(in_features=768, out_features=768, bias=False)
          (k): Linear(in_features=768, out_features=768, bias=False)
          (v): Linear(in_features=768, out_features=768, bias=False)
          (o): Linear(in_features=768, out_features=768, bias=False)
        )
        (layer_norm): T5LayerNorm()
        (dropout): Dropout(p=0.1, inplace=False)
      )
      (2): T5LayerFF(
        (DenseReluDense): T5DenseActDense(
          (wi): Linear(in_features=768, out_features=3072, bias=False)
          (wo): Linear(in_features=3072, out_features=768, bias=False)
          (dropout): Dropout(p=0.1, inplace=False)
          (act): ReLU()
        )
        (layer_norm): T5LayerNorm()
        (dropout): Dropout(p=0.1, inplace=False)
      )
    )
  )
  (7): T5Block(
    (layer): ModuleList(
      (0): T5LayerSelfAttention(
        (SelfAttention): T5Attention(
          (q): Linear(in_features=768, out_features=768, bias=False)
          (k): Linear(in_features=768, out_features=768, bias=False)
          (v): Linear(in_features=768, out_features=768, bias=False)
          (o): Linear(in_features=768, out_features=768, bias=False)
        )
        (layer_norm): T5LayerNorm()
        (dropout): Dropout(p=0.1, inplace=False)
      )
      (1): T5LayerCrossAttention(
        (EncDecAttention): T5Attention(
          (q): Linear(in_features=768, out_features=768, bias=False)
          (k): Linear(in_features=768, out_features=768, bias=False)
          (v): Linear(in_features=768, out_features=768, bias=False)
          (o): Linear(in_features=768, out_features=768, bias=False)
```

```
          )
          (layer_norm): T5LayerNorm()
          (dropout): Dropout(p=0.1, inplace=False)
        )
        (2): T5LayerFF(
          (DenseReluDense): T5DenseActDense(
            (wi): Linear(in_features=768, out_features=3072, bias=False)
            (wo): Linear(in_features=3072, out_features=768, bias=False)
            (dropout): Dropout(p=0.1, inplace=False)
            (act): ReLU()
          )
          (layer_norm): T5LayerNorm()
          (dropout): Dropout(p=0.1, inplace=False)
        )
      )
    )
    (8): T5Block(
      (layer): ModuleList(
        (0): T5LayerSelfAttention(
          (SelfAttention): T5Attention(
            (q): Linear(in_features=768, out_features=768, bias=False)
            (k): Linear(in_features=768, out_features=768, bias=False)
            (v): Linear(in_features=768, out_features=768, bias=False)
            (o): Linear(in_features=768, out_features=768, bias=False)
          )
          (layer_norm): T5LayerNorm()
          (dropout): Dropout(p=0.1, inplace=False)
        )
        (1): T5LayerCrossAttention(
          (EncDecAttention): T5Attention(
            (q): Linear(in_features=768, out_features=768, bias=False)
            (k): Linear(in_features=768, out_features=768, bias=False)
            (v): Linear(in_features=768, out_features=768, bias=False)
            (o): Linear(in_features=768, out_features=768, bias=False)
          )
          (layer_norm): T5LayerNorm()
          (dropout): Dropout(p=0.1, inplace=False)
        )
        (2): T5LayerFF(
          (DenseReluDense): T5DenseActDense(
            (wi): Linear(in_features=768, out_features=3072, bias=False)
            (wo): Linear(in_features=3072, out_features=768, bias=False)
            (dropout): Dropout(p=0.1, inplace=False)
            (act): ReLU()
          )
          (layer_norm): T5LayerNorm()
          (dropout): Dropout(p=0.1, inplace=False)
        )
      )
    )
    (9): T5Block(
      (layer): ModuleList(
        (0): T5LayerSelfAttention(
          (SelfAttention): T5Attention(
            (q): Linear(in_features=768, out_features=768, bias=False)
            (k): Linear(in_features=768, out_features=768, bias=False)
            (v): Linear(in_features=768, out_features=768, bias=False)
            (o): Linear(in_features=768, out_features=768, bias=False)
          )
          (layer_norm): T5LayerNorm()
          (dropout): Dropout(p=0.1, inplace=False)
        )
```

```
      (1): T5LayerCrossAttention(
        (EncDecAttention): T5Attention(
          (q): Linear(in_features=768, out_features=768, bias=False)
          (k): Linear(in_features=768, out_features=768, bias=False)
          (v): Linear(in_features=768, out_features=768, bias=False)
          (o): Linear(in_features=768, out_features=768, bias=False)
        )
        (layer_norm): T5LayerNorm()
        (dropout): Dropout(p=0.1, inplace=False)
      )
      (2): T5LayerFF(
        (DenseReluDense): T5DenseActDense(
          (wi): Linear(in_features=768, out_features=3072, bias=False)
          (wo): Linear(in_features=3072, out_features=768, bias=False)
          (dropout): Dropout(p=0.1, inplace=False)
          (act): ReLU()
        )
        (layer_norm): T5LayerNorm()
        (dropout): Dropout(p=0.1, inplace=False)
      )
    )
  )
  (10): T5Block(
    (layer): ModuleList(
      (0): T5LayerSelfAttention(
        (SelfAttention): T5Attention(
          (q): Linear(in_features=768, out_features=768, bias=False)
          (k): Linear(in_features=768, out_features=768, bias=False)
          (v): Linear(in_features=768, out_features=768, bias=False)
          (o): Linear(in_features=768, out_features=768, bias=False)
        )
        (layer_norm): T5LayerNorm()
        (dropout): Dropout(p=0.1, inplace=False)
      )
      (1): T5LayerCrossAttention(
        (EncDecAttention): T5Attention(
          (q): Linear(in_features=768, out_features=768, bias=False)
          (k): Linear(in_features=768, out_features=768, bias=False)
          (v): Linear(in_features=768, out_features=768, bias=False)
          (o): Linear(in_features=768, out_features=768, bias=False)
        )
        (layer_norm): T5LayerNorm()
        (dropout): Dropout(p=0.1, inplace=False)
      )
      (2): T5LayerFF(
        (DenseReluDense): T5DenseActDense(
          (wi): Linear(in_features=768, out_features=3072, bias=False)
          (wo): Linear(in_features=3072, out_features=768, bias=False)
          (dropout): Dropout(p=0.1, inplace=False)
          (act): ReLU()
        )
        (layer_norm): T5LayerNorm()
        (dropout): Dropout(p=0.1, inplace=False)
      )
    )
  )
  (11): T5Block(
    (layer): ModuleList(
      (0): T5LayerSelfAttention(
        (SelfAttention): T5Attention(
          (q): Linear(in_features=768, out_features=768, bias=False)
          (k): Linear(in_features=768, out_features=768, bias=False)
```

```
              (v): Linear(in_features=768, out_features=768, bias=False)
              (o): Linear(in_features=768, out_features=768, bias=False)
            )
            (layer_norm): T5LayerNorm()
            (dropout): Dropout(p=0.1, inplace=False)
          )
          (1): T5LayerCrossAttention(
            (EncDecAttention): T5Attention(
              (q): Linear(in_features=768, out_features=768, bias=False)
              (k): Linear(in_features=768, out_features=768, bias=False)
              (v): Linear(in_features=768, out_features=768, bias=False)
              (o): Linear(in_features=768, out_features=768, bias=False)
            )
            (layer_norm): T5LayerNorm()
            (dropout): Dropout(p=0.1, inplace=False)
          )
          (2): T5LayerFF(
            (DenseReluDense): T5DenseActDense(
              (wi): Linear(in_features=768, out_features=3072, bias=False)
              (wo): Linear(in_features=3072, out_features=768, bias=False)
              (dropout): Dropout(p=0.1, inplace=False)
              (act): ReLU()
            )
            (layer_norm): T5LayerNorm()
            (dropout): Dropout(p=0.1, inplace=False)
          )
        )
      )
    )
    (final_layer_norm): T5LayerNorm()
    (dropout): Dropout(p=0.1, inplace=False)
  )
  (lm_head): Linear(in_features=768, out_features=32100, bias=False)
)
```